\documentclass[letterpaper, paper,11pt]{AAS}        

\usepackage{bm}
\usepackage{amsmath}
\usepackage{subfigure}
\usepackage{overcite}
\usepackage{footnpag}                   

\usepackage[ruled,vlined,linesnumbered]{algorithm2e}

\usepackage{amssymb}

\usepackage{url}

\PaperNumber{19-350}

\begin{document}

\title{Pose Estimation For Non-Cooperative Spacecraft Rendezvous Using Neural Networks}

\author{Sumant Sharma\thanks{Ph.D. Candidate, Department of Aeronautics \& Astronautics, Stanford University, Stanford, California, USA.},  
Simone D'Amico\thanks{Assistant Professor, Department of Aeronautics \& Astronautics, Stanford University, Stanford, California, USA.}
}

\maketitle{}

\begin{abstract}
This work introduces the Spacecraft Pose Network (SPN) for on-board estimation of the pose, i.e., the relative position and attitude, of a known non-cooperative spacecraft using monocular vision. In contrast to other state-of-the-art pose estimation approaches for spaceborne applications, the SPN method does not require the formulation of hand-engineered features and only requires a single grayscale image to determine the pose of the spacecraft relative to the camera. The SPN method uses a Convolutional Neural Network (CNN) with three branches to solve the problems of relative attitude and relative position estimation. The first branch of the CNN bootstraps a state-of-the-art object detection algorithm to detect a 2D bounding box around the target spacecraft in the input image. The region inside the 2D bounding box is then used by the other two branches of the CNN to determine the relative attitude by initially classifying the input region into discrete coarse attitude labels before regressing to a finer estimate. The SPN method then uses a novel Gauss-Newton algorithm to estimate the relative position by using the constraints imposed by the detected 2D bounding box and the estimated relative attitude. The secondary contribution of this work is the generation of the Spacecraft PosE Estimation Dataset (SPEED), which is used to train and evaluate the performance of the SPN method. SPEED consists of synthetic as well as actual camera images of a mock-up of the Tango spacecraft from the PRISMA mission. The synthetic images are created by fusing OpenGL-based renderings of the spacecraft's 3D model with actual images of the Earth captured by the Himawari-8 meteorological satellite. The actual camera images are created using a 7 degrees-of-freedom robotic arm, which positions and orients a vision-based sensor with respect to a full-scale mock-up of the Tango spacecraft. Custom illumination devices simulate the Earth albedo and Sun light with high fidelity to emulate the illumination conditions present in space. The SPN method, trained only on synthetic images, produces degree-level relative attitude error and cm-level relative position errors when evaluated on the actual camera images not used during training.

\end{abstract}

\section{Introduction}

The on-board estimation of the pose, i.e., the relative position and attitude, of a known non-cooperative spacecraft using a monocular camera is a key-enabling technology for current and future on-orbiting servicing and debris removal missions such as the RemoveDEBRIS mission by Surrey Space Centre\cite{Forshaw2016}, the Phoenix program by DARPA \cite{Sullivan2013}, and the Restore-L mission by NASA\cite{Reed2016}. The knowledge of the current pose of the target spacecraft during the proximity operations of these missions allows real-time approach trajectory guidance and control. In contrast to systems based on LiDAR and stereo camera sensors, monocular navigation ensures pose estimation under low mass and power requirements, making it a natural sensor candidate for future formation-flying missions especially employing small satellites.

Prior demonstrations of close-range pose estimation have utilized image processing based on hand-engineered features\cite{DAmico2013,Sharma2018, Cropp2002,Liu2014,Naasz2010,Capuano2018,Kelsey2006,Sharma2015c,Lunghi2018} and an a-priori knowledge of the pose\cite{Petit2011,Zhang2005,Aviles2016,SharmaAasGnc2017}. A key strength for many of these methods is their use of the perspective transformation between the scene and the image to hypothesize and test feature correspondences detected in the 2D image and the known 3D model of target spacecraft. However, the formulation of specific features is not scalable to spacecraft of different structural and physical properties as well as not robust to the dynamic illumination conditions of space. Secondly, the a-priori knowledge of the pose of the target is not always available due to mission operations constraints nor desirable when full autonomy is required. 

Recent advancements in pose estimation techniques for terrestrial applications have relied on deep learning algorithms. Broadly, these algorithms bypass the classical image processing based pipeline and instead attempt to learn the non-linear transformation between the two-dimensional input image space and the six-dimensional output pose space in an end-to-end fashion. The deep learning-based methods either discretize the pose space and solve the resulting classification problem \cite{Sharma2018a,Su2016,Tulsiani2015,Sharma2017} or directly regress the relative pose from the input image \cite{Mahendran2018,Kendall2015,Xiang2017}. Render-for-CNN \cite{Su2016} demonstrated the use of rendered images to train a convolutional neural network for viewpoint estimation on actual camera images. The viewpoint estimation problem is casted as classifying the camera rotation parameters into fine-grained bins. PoseCNN \cite{Xiang2017} used separate branches of a convolutional neural network to predict semantic labels, object centers in the 2D image, and object rotation using direct regression. However, the approach is not accurate enough without further refinement using the iterative closest point method\cite{Besl1992}. Classification-based approaches rely on the fine discretization of the pose space into a large number of pose labels in order to achieve reasonable pose accuracy. On the other hand, direct regression-based approaches require careful choice of parameters to avoid unpredictable behavior while learning the transformation between the input two-dimensional pixel information and the output regression parameters describing the six-dimensional pose space. In addition, the applicability of these algorithms to space imagery is not trivial due to two reasons. Firstly, unlike terrestrial applications, spaceborne navigation cameras are challenged with quickly varying illumination conditions and capture imagery with low signal-to-noise ratio and high contrast. Secondly, the massive datasets necessary to train machine learning algorithms are typically not available for spaceborne navigation.

\begin{figure}[htbp]
    \centering
    \includegraphics[width=0.99\textwidth]{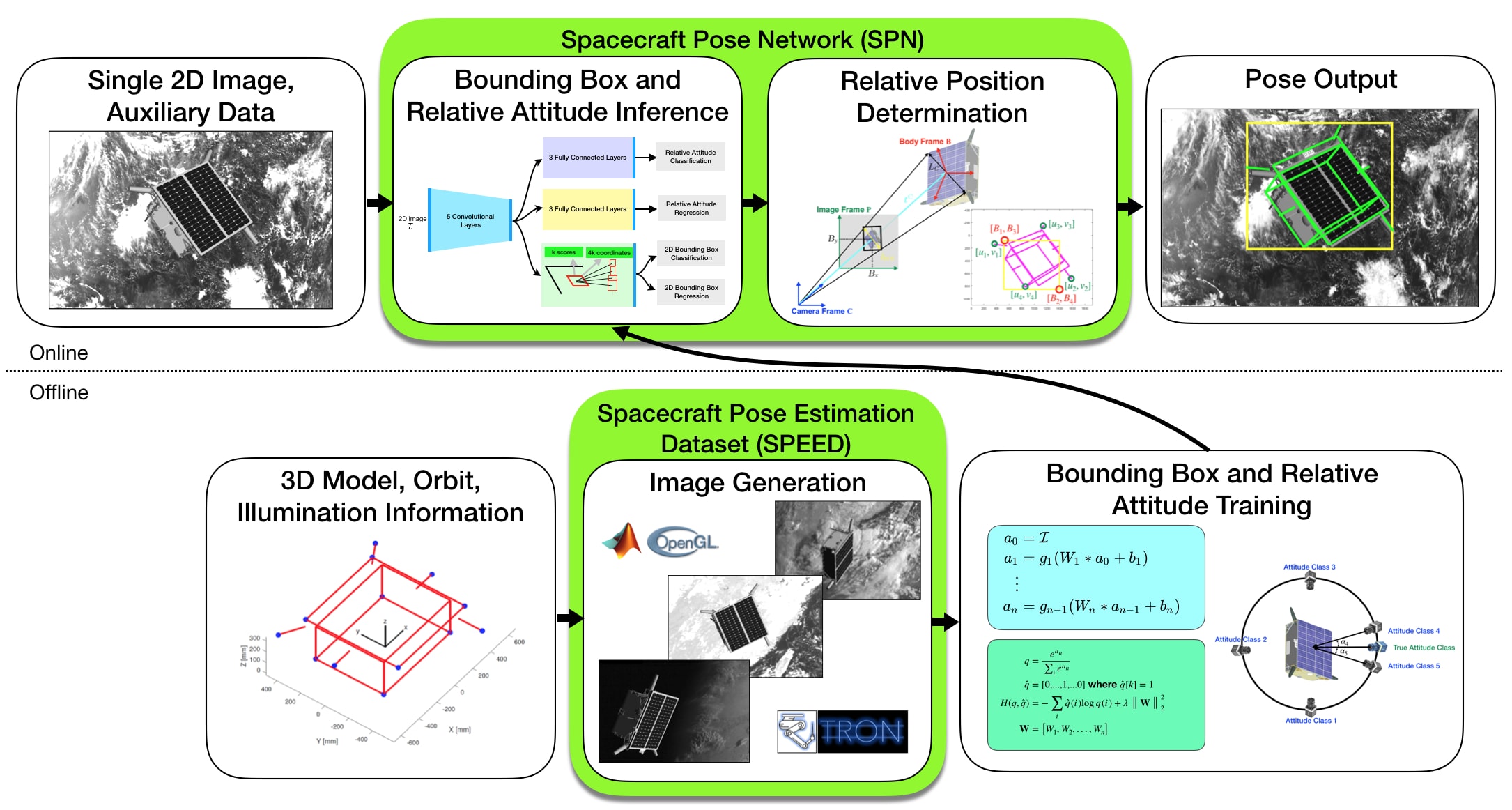}
    \caption{Modules of the proposed SPN method, which takes as input a 2D image and a trained convolutional neural network for relative attitude and position determination.}
    \label{fig:meas}
\end{figure}
The primary contribution of this work is the introduction of the Spacecraft Pose Network (SPN), a new deep learning-based pose estimation method. As shown in Figure~\ref{fig:meas}, the SPN method uses a convolutional neural network to estimate the relative position and relative attitude in a decoupled fashion from a single grayscale image. One branch of the convolutional neural network is used to detect a 2D bounding box around the target spacecraft in the input image. The other two branches are used to estimate the relative attitude using a hybrid discrete-continuous method. The relative attitude and the 2D bounding box are then combined with geometrical constraints of the perspective transformation to estimate the relative position using the Gauss-Newton algorithm. In contrast to current deep learning-based techniques, the relative attitude accuracy provided by the SPN method is not limited by the level of discretization of the pose space and it explicitly uses the geometrical knowledge of the perspective transformation in the estimation of relative position. As compared with techniques comparing image features with features of a known target spacecraft model on-board, the SPN method provides a pose estimate from a single image without requiring a long initialization phase or favorable relative translation motion. Further, due to the decoupling of the relative attitude and position estimation and the use of transfer learning, the SPN method can be used to infer the pose of the target spacecraft through training on a relatively small number of synthetic images of the same spacecraft.

The secondary contribution of this work is the creation of the Spacecraft Pose Estimation Dataset (SPEED), which  consists of high-fidelity imagery involving close proximity operations around a tumbling spacecraft. The dataset contains 15,300 images from two sources: a purely software-based Augmented Reality (AR) source that fuses synthetic and actual space imagery, and a purely reality-based source that uses an actual camera sensor to capture images of a mock-up spacecraft under high-fidelity illumination conditions. The convolutional neural network of the SPN method is trained only on a portion of these AR images while it is tested on the remaining AR images as well as the actual camera images of SPEED. SPEED will also be made publicly available to the community to enable a fair comparison of the state-of-the-art pose estimation techniques through a competition on pose estimation for spaceborne applications organized in collaboration with the European Space Agency\cite{EuropeanSpaceAgencyURL}. 

The paper is organized as follows: Section 2 describes the framework for the SPN method; Section 3 describes the SPEED image generation; Section 4 describes the experiments conducted to validate the performance of SPN method; and Section 5 presents conclusions from this work as well as directions for further research and development.

\newpage
\section{Spacecraft Pose Network}

\begin{figure}[htbp]
\includegraphics[width=2.5in]{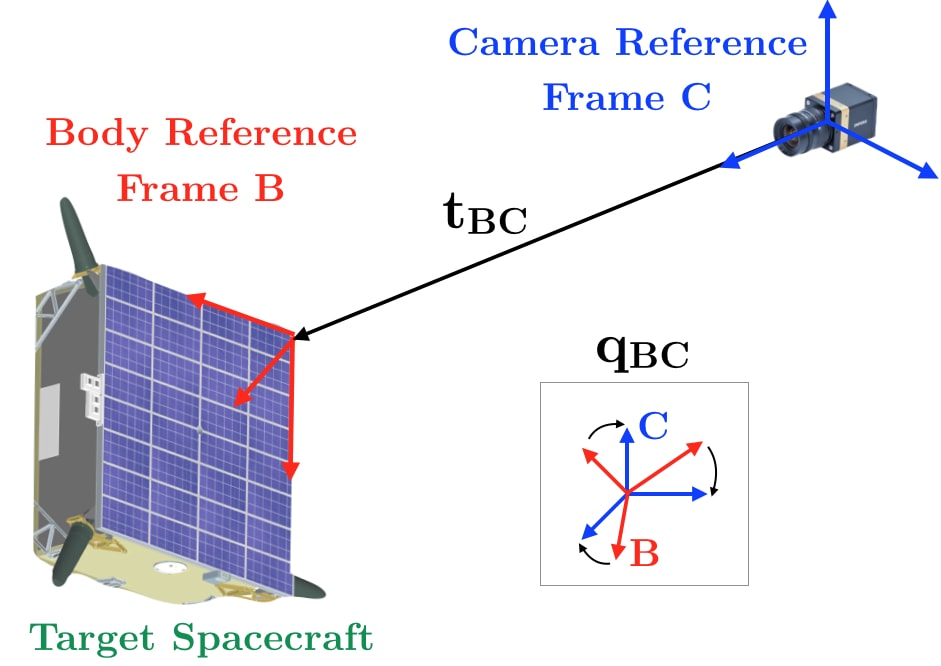}
\centering
\caption{Definition of the reference frames, relative position, and relative attitude.}
\label{fig:posedet}
\end{figure}

Formally, the problem statement for this work is the estimation of the attitude and position of the camera frame, $\text{C}$, with respect to the body frame of the target spacecraft, $\text{B}$. As shown in Figure~\ref{fig:posedet}, $\mathbf{\mathbf{t_{BC}}}$ is the relative position of the origin of the target's body reference frame w.r.t. the origin of the camera's reference frame. Similarly, $\mathbf{\mathbf{q_{BC}}}$ is the quaternion associated with the rotation that aligns the target's body reference frame with the camera's reference frame.

\begin{figure}[htbp]
\includegraphics[width=0.99\textwidth]{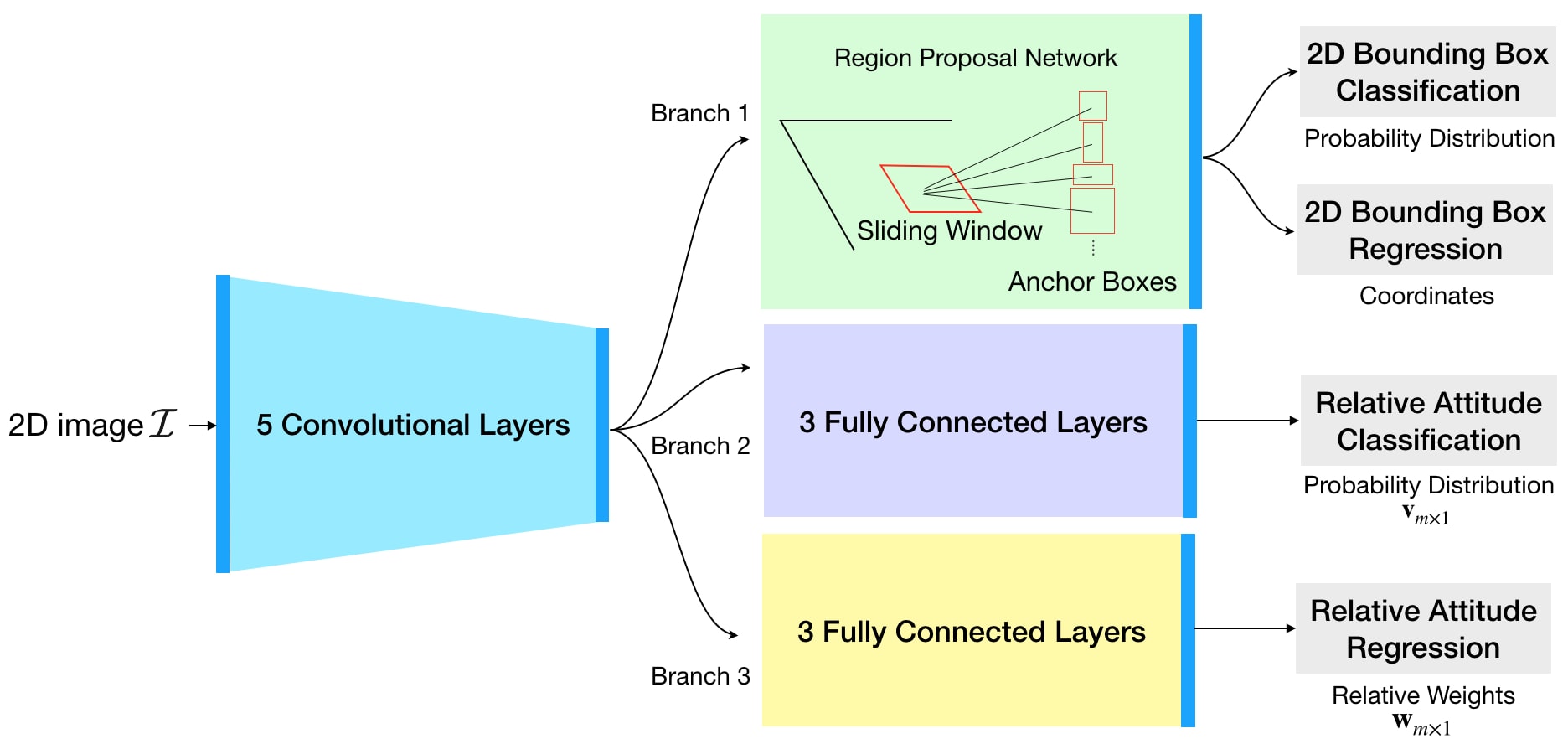}
\centering
\caption{Illustration of the convolutional neural network used in the SPN method. Branch 1 uses the region proposal network \cite{Ren2015} to detect a 2D bounding box around the target spacecraft while Branches 2 and 3 of the network are used in a hybrid classification-regression fashion to obtain the relative position.}
\label{fig:spn}
\end{figure}

As shown in Figure~\ref{fig:spn}, the SPN method uses three separate branches of a convolutional neural network to estimate $\mathbf{t_{BC}}$ and $\mathbf{q_{BC}}$. Branch 1 is used to detect a 2D bounding box in the image around the target spacecraft. The output of the convolutional layers corresponding to the detected 2D bounding box is used as an input to Branches 2 and 3 to obtain an estimate of the relative attitude, $\mathbf{\tilde{q}_{BC}}$. Finally, the 2D bounding box and $\mathbf{\tilde{q}_{BC}}$ are input to the Gauss-Newton algorithm to estimate the relative position as $\mathbf{\tilde{t}_{BC}}$. The estimation of the relative position and relative attitude are discussed in detail in the following subsections.

\subsection{Relative Position Estimation}
The relative position estimation begins with the detection of a 2D bounding box in the image around the target spacecraft using the region proposal network \cite{Ren2015}. The detected 2D bounding box is not only important to effectively remove the background from the image before relative attitude estimation but makes the SPN method extensible to perform pose estimation for multiple objects or spacecraft components from the same image. The region proposal network used in the SPN method is an ``off-the-shelf'' object detection algorithm that takes the output of the five convolutional layers as an input. This network uses a sliding window approach on the output of the convolutional layers to produce region proposals based on predefined anchor boxes. Each anchor box is centered at the sliding window in question, and is associated with a scale and aspect ratio. Typically, 3 scales and 3 aspect ratios are used, yielding $k$ = 9 anchor boxes at each sliding position. Therefore, for each sliding window location, the region proposal network outputs $k$ probabilities (or scores) of whether the target spacecraft is present and regresses to $4k$ coordinates of the corresponding $k$ 2D bounding boxes. Finally, the 2D bounding box associated with the highest probability is provided as the output. The reader can find further implementation details in the original paper \cite{Ren2015}. Even though the SPN method uses the region proposal network in its current implementation, it can be easily swapped with another state-of-the-art object detection algorithm \cite{Kaushal2018} based on specific computation runtime, storage, and accuracy requirements.

The resulting 2D bounding box estimated by the region proposal network and the relative attitude estimated by Branches 2 and 3 of the SPN method are combined with geometrical constraints to estimate the relative position using the Gauss-Newton algorithm. The SPN method uses the fact that the perspective projection of the 3D wireframe model of the target spacecraft must fit tightly within the detected 2D bounding box. 
\begin{figure}[htbp]
    \centering
    \includegraphics[width=0.3\textwidth]{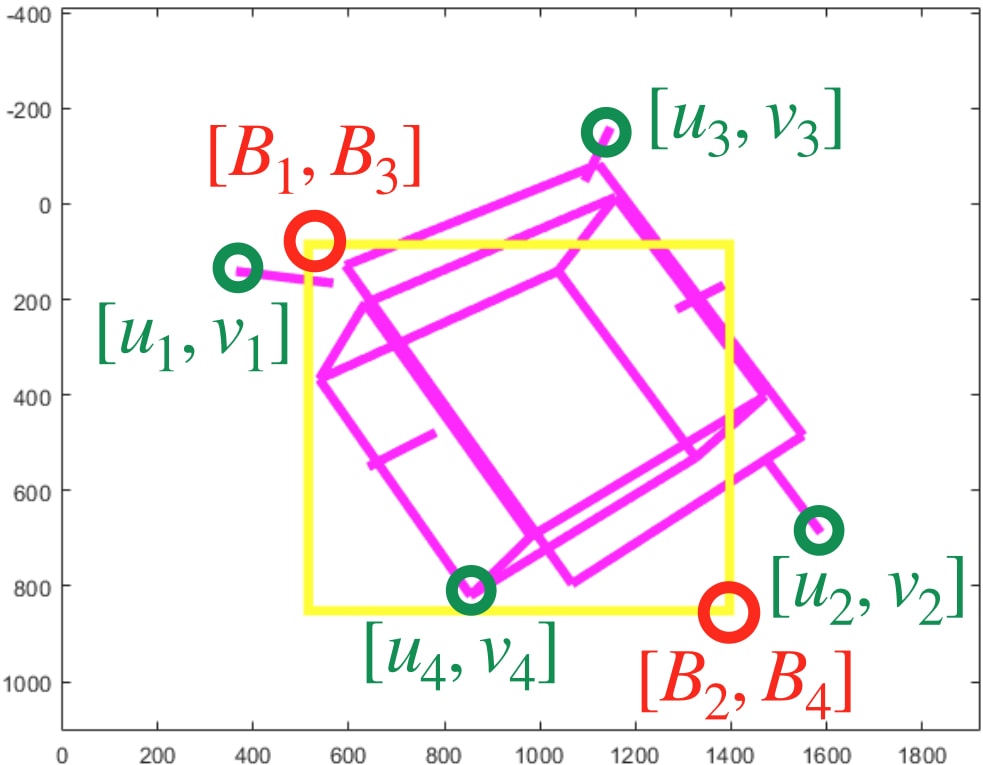}
    \caption{Schematic of the projection of the 3D wireframe model of the target (pink) and the estimated bounding box (yellow) in the image plane.}
    \label{fig:relposdet}
\end{figure}

As shown in Figure~\ref{fig:relposdet}, the 2D bounding box can be parametrized by its top-left coordinates, $(B_1, B_3)$, and its bottom-right coordinates, $(B_2, B_4)$, defined in the image plane. Given the estimated relative attitude parametrized as a rotation matrix, $R(\mathbf{\tilde{q}_{BC}})$, the camera focal lengths, $(f_x, f_y)$, and the camera principal points, $(c_x, c_y)$, the 3D points defined in the body frame of the target spacecraft, $\mathbf{X}_i$, can be projected into the image plane using the perspective equation,
\begin{equation}
\label{eq:perspective}
\begin{bmatrix}
u_i w_i\\ 
v_i w_i\\ 
w_i
\end{bmatrix} = 
\begin{bmatrix}
f_x & 0 & c_x\\ 
0 & f_y & c_y\\ 
0 & 0 & 1
\end{bmatrix}
\begin{bmatrix} R(\mathbf{\tilde{q}_{BC}}) & \mathbf{t_{BC}} \end{bmatrix} \mathbf{X}_i.
\end{equation}
The constraint that the perspective projection of the 3D wireframe model fits tightly within the 2D bounding box requires that the four extremal projected points, $(u_i, v_i)$, be as close as possible to the four edges of the 2D bounding box. Specifically, the four extremal projected points are the ``left-most'', ``right-most'', ``top-most'', and ``bottom-most'' points, respectively. For example, $(u_1, v_1)$ is the ``left-most'' projected point and it is constrained to be as close as possible to $B_1$, the coordinate representing the left edge of the 2D bounding box. Mathematically, the SPN method solves the following minimization problem subject to the constraint posed by Equation~\ref{eq:perspective}
\begin{equation}
\begin{aligned}
& \underset{\mathbf{t_{BC}}}{\text{minimize}}
& & \sum\limits_{i=1}^2 (u_i(\mathbf{t_{BC}}) - B_{i})^2 + \sum\limits_{i=3}^4 (v_i(\mathbf{t_{BC}}) - B_{i})^2.
\end{aligned}
\end{equation}

The minimization problem is solved using the Gauss-Newton algorithm, which requires an initial guess of $\mathbf{t_{BC}}$. The SPN method uses the detected 2D bounding box and the characteristic length of the 3D wireframe model to provide this initial guess. 
\begin{figure}[!ht]
  \centering
    \includegraphics[width=0.5\textwidth]{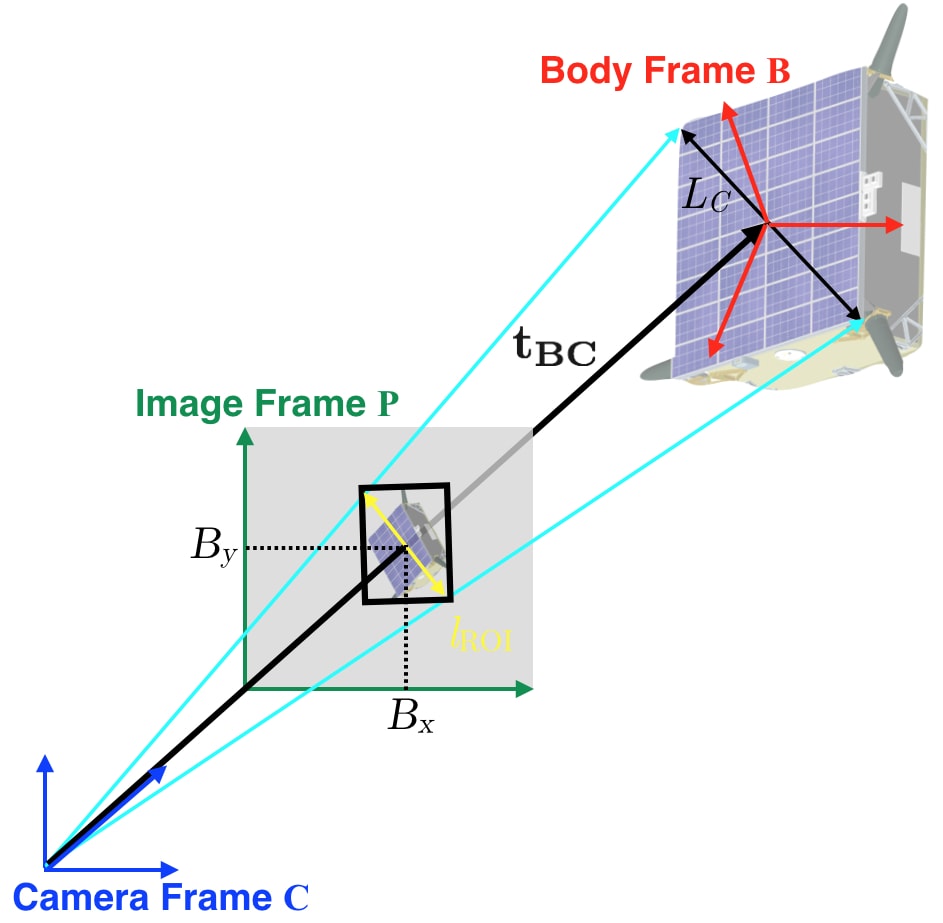}
    \caption{Calculation of the relative position using the 2D bounding box.} \label{fig:wgePosition}
\end{figure}

Figure \ref{fig:wgePosition} shows that the knowledge of the diagonal characteristic length, $L_C$, of the spacecraft 3D model and the diagonal length of the detected 2D bounding box, $l_{\text{ROI}}$, can be used to obtain $\mathbf{t_{BC}}$. In particular, a coarse estimate of the distance to the target spacecraft from the origin of the camera frame is
\begin{equation}
||\mathbf{t_{BC}}||_2 = \frac{\left(\frac{f_x + f_y}{2}\right)L_C}{l_\text{ROI}}.
\end{equation}
Assuming that the origin of the target spacecraft body frame, $\text{B}$, lies on the ray projected from the origin of the camera frame, $\text{C}$, towards the center of the detected 2D bounding box, $(B_x, B_y)$, azimuth and elevation angles, $(\alpha,\beta)$, can be derived using $(c_x, c_y)$ and $(B_x,B_y)$
\begin{eqnarray}
\alpha = \tan^{-1}\left(\frac{B_x - c_x}{f_x} \right) \label{eq:bearingAngles1}\\
\beta = \tan^{-1}\left(\frac{B_y - c_y}{f_y} \right).
\label{eq:bearingAngles2}
\end{eqnarray}
Finally, the coarse relative position used as the initial guess in the Gauss-Newton algorithm is
\begin{equation}
\mathbf{t_{BC}} = \begin{bmatrix}
\cos(\alpha) & 0 & -\sin(\alpha)\\ 
0 & 1 & 0 \\ 
\sin(\alpha) & 0 & \cos(\alpha) 
\end{bmatrix} 
\begin{bmatrix}
1 & 0 & 0\\ 
0 & \cos(\beta) & \sin(\beta )\\ 
0 & -\sin(\beta )& \cos(\beta
)\end{bmatrix}
\begin{bmatrix}
0\\ 
0\\ 
||\mathbf{t_{BC}}||_2
\end{bmatrix}.
\end{equation}

\subsection{Relative Attitude Determination}
\begin{figure}[h!]
\includegraphics[width=0.5\linewidth]{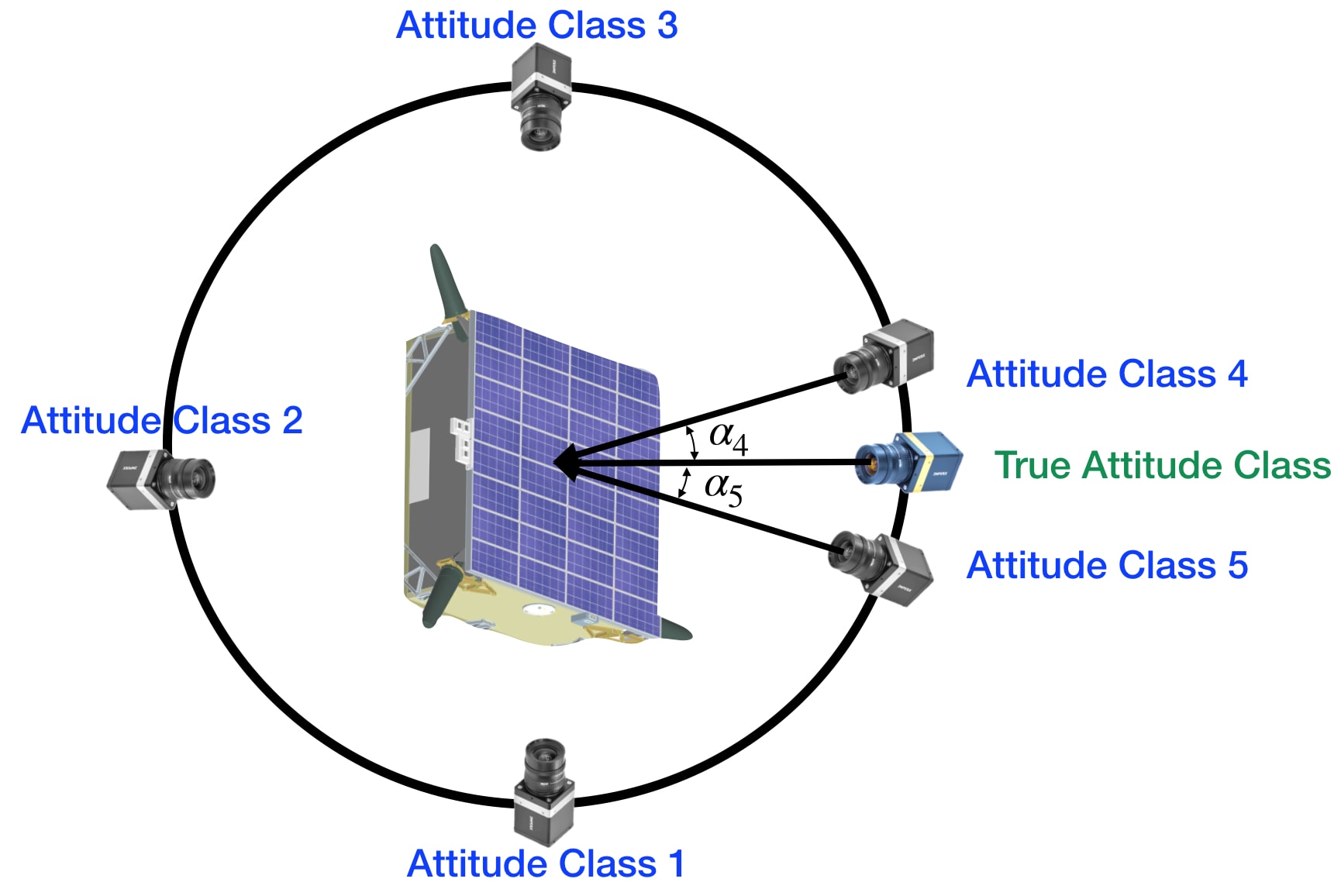}
\centering
\caption{Visualization of the relative attitude discretization in a single dimension. Here attitude classes 4 and 5 are adjacent classes closest to the ground truth attitude.}
\label{fig:attDisc}
\end{figure}

The layers of a convolutional neural network transform the raw pixel information, $\mathcal{I}$, to gradually more abstract representations using predefined non-linear functions that contain unknown coefficients. The output of the network is then constrained to minimize a loss function that represents a discrepancy or error between the output of the network and the expected output for a set of training samples (supervised learning). For example, the network used in the SPN method can be represented by
\begin{equation}
\begin{split}
a_{0} &= \mathcal{I} \\
a_{1} &= g_1(W_1 * a_0 + b_1) \\
&\vdots \\
a_{n} &= g_{n-1}(W_n * a_{n-1} + b_n)
\end{split}    
\end{equation}
where $W_i$ are the unknown weights for the $i$-th layer, $b_i$ are the unknown biases for the $i$-th layer, and $g_i$ are the known non-linear functions for the $i$-th layer. For relative attitude estimation, the network's output is expected to be values defined on a continuous non-Euclidean space, prohibiting the direct use of a typical L2 loss function. To handle this problem, other authors have proposed several loss functions \cite{Xiang2017,Hara2017,Wu2018} that directly regress to a relative attitude estimate, however, these have limited accuracy and require further refinement \cite{Li}. Instead, the SPN method uses two branches of fully-connected layers that share the output of five preceding convolutional layers as their inputs. In order to be robust to the background in the images, the features output by the final convolutional layer associated with the detected 2D bounding box are selected using the RoI pooling layer\cite{Girshick2015}. Using these features as input, Branch 2 performs a classification task to find the closest predefined attitude classes that describe the input image. Branch 3 uses these features as input to perform a regression task to find the relative weights of the attitude classes identified in Branch 2. For example, Figure~\ref{fig:attDisc} shows a simplified example where one dimension of the relative attitude has been discretized. For this case, Branch 2 is expected to find attitude classes 4 and 5 as the two adjacent classes that best describe the input image while Branch 3 is expected to find relative weights as a function of the angular differences $\alpha_4$ and $\alpha_5$ associated with the corresponding attitude classes. The sizes of the five convolutional layers and both sets of the three fully connected layers have been adopted from the Zeiler and Fergus model architecture \cite{Zeiler2013}.

The attitude classes are parametrized as $m$ unit quaternions representing uniformly distributed random rotations in the $SO(3)$ space. Algorithm~\ref{alg:subgroup} shows the subgroup algorithm \cite{Shoemake1992} used to obtain these random rotations. The algorithm amounts to multiplying a uniformly distributed element from the subgroup of planar rotations with a uniformly distributed coset represented by the rotations pointing the z-axis in different directions. 

\begin{algorithm}[htbp]
\caption{Computes $m$ uniformly distributed random rotations parametrized as unit quaternions.}
\label{alg:subgroup}
\DontPrintSemicolon
\SetAlgoLined
\SetKwInOut{Input}{Input}\SetKwInOut{Output}{Output}
\Input{Integer $m$}
\Output{Matrix quats}
\BlankLine
$x_0\leftarrow$ rand($m$) \texttt{// get m uniformly distributed float values $\in$ [0,1]}\;
$x_1\leftarrow$ rand($m$)\;
$x_2\leftarrow$ rand($m$)\;
$\theta_1 = 2 * \pi * x1$\;
$\theta_2 = 2 * \pi * x2$\;
\BlankLine
$s_1 = \sin(\theta_1)$\;
$s_2 = \sin(\theta_2)$\;
$c_1 = \cos(\theta_1)$\;
$c_2 = \cos(\theta_2)$\;
\BlankLine
$r_1 = \sqrt{1 - x_0}$\;
$r_2 = \sqrt{x_0}$\;
\BlankLine
quats = [$s_1$ .* $r_1$, $c_1$ .* $r_1$, $s_2$ .* $r_2$, $c_2$ .* $r_2$]\; 
\end{algorithm}

To determine which of the $m$ predefined attitude classes are the closest to the given image, Branch 2 is tasked to output an $m \times 1$ vector $\mathbf{v}$, which represents a probability distribution. Each entry $\mathbf{v}_j$ is the probability whether the attitude class in question is one of the $n$ closest attitude classes to the relative attitude of the given image. Note that both $m$ and $n$ are hyper-parameters. Closeness is defined by the angular difference between the unit quaternion representing the attitude class, $\mathbf{q}_i$, and the unit quaternion representing the ground truth value of the relative attitude, $\mathbf{q_{BC}}$. To ensure that $\mathbf{v}$ is a valid probability distribution, the output of the final layer of the Branch 2 is passed through the softmax function
\begin{equation}
    \sigma(\mathbf{v})_j = \frac{e^{\mathbf{v}_j}}{\sum\limits_{k=1}^m e^{\mathbf{v}_j}} \;\;\; \forall j \in [1,m]
\end{equation}
Then the loss function for Branch 2, $L_\text{class}$, representing the difference between the branch output and the expected probability distribution, $\tilde{\mathbf{v}}$, can be written as
\begin{equation}
    L_\text{class} = - \sum\limits_{j=1}^m \tilde{\mathbf{v}}_j \log (\sigma(\mathbf{v})_j) + \lambda || \mathbf{W_1} ||_2^2,
\end{equation}
where $\mathbf{W_1}$ represents the weights of the final three layers of Branch 2 and $\lambda$ is a scalar representing the strength of the L2 regularization. The aim of the L2 regularization is to penalize the existence of large weights and prevent overfitting to the training examples. Note that the entries of $\tilde{\mathbf{v}}$ corresponding to the $n$ closest attitude classes is set to $1/n$ while the rest of the entries are set to zero.

To estimate the relative attitude for the given image, Branch 3 is tasked to output an $m \times 1$ vector, $\mathbf{w}$, containing weights for the $n$ closest attitude classes identified in Branch 2. The weights for the remaining $m-n$ attitude classes are output but not used during either training or inference. The weights of the $n$ closest attitude classes are passed through the softmax function such that their sum adds to unity. Let $\Omega$ be the vector of indices of the $n$ largest values in $\sigma(\mathbf{v})$, then the loss function for the second branch can be written as
\begin{equation}
    L_\text{reg} = - \sum\limits_{j=1}^m \tilde{\mathbf{w}}_j \log \left (\frac{e^{\mathbf{w}_j}}{\sum_j e^{\mathbf{w}_j}} \right ) + \lambda || \mathbf{\mathbf{W}_2}||_2^2 \;\;\; \forall j \in \Omega,
\end{equation}
where $\mathbf{W_2}$ represents the weights of the final three layers of Branch 3 and $\tilde{\mathbf{w}}$ is the ground truth vector of weights for the given image. The entries of $\tilde{\mathbf{w}}$ are set based on the angular difference between the quaternion of the attitude class in question and the ground truth quaternion of the given image. Specifically,
\begin{equation}
    \tilde{\mathbf{w}}_j = \frac{1 - \bm{\alpha}_j / \pi^2}{n - \sum_j \bm{\alpha}_j / \pi^2 }, \forall j \in \Omega,
\end{equation}
where $\bm{\alpha}_j$ is the angular difference between the unit quaternion representing the attitude class in question and the unit quaternion representing the ground truth of the relative attitude of the given image. Hence, the total loss function for relative attitude estimation can be written as:
\begin{equation}
    L_\text{total} = L_\text{class} + \mu L_\text{reg}
\end{equation}

The classification loss, $L_\text{class}$, is designed to force Branch 2's coefficients to be correlated with macroscopic changes in the relative attitude without penalizing classification predictions of attitude classes that have small angular difference between them. Instead, the regularization loss, $L_\text{reg}$ is designed to force the Branch 3's coefficients to be correlated with the microscopic differences between adjacent attitude classes.

During inference, the estimate of the relative attitude, $\mathbf{\tilde{q}_{BC}}$, is computed using an average of the unit quaternions of the $n$ closest attitude classes, $\mathbf{Q}$, weighted by their corresponding relative weights, $\mathbf{\Gamma}$. This is akin to minimizing a weighted sum of the squared Frobenius norms of the differences between the rotation matrix representation of $\mathbf{Q}_i$ and $\mathbf{q}$
\begin{equation}
    \mathbf{\tilde{q}_{BC}} = \arg \underset{\mathbf{q} \in \mathbb{S}^3}{\min} \sum\limits_{i} \mathbf{\Gamma}_i || R(\mathbf{q}) - R(\mathbf{Q}_i) ||^2_F
\end{equation}
where $\mathbb{S}^3$ denotes the unit 3-sphere\cite{Markley2007}. The pseudo-code used to compute the weighted average is presented below as Algorithm~\ref{alg:avg}.

\begin{algorithm}[htbp]
\caption{Computes a unit quaternion that is an average of the input unit quaternions $\mathbf{Q}$ weighted by the input weights $\mathbf{\Gamma}$.}
\label{alg:avg}
\DontPrintSemicolon
\SetAlgoLined
\SetKwInOut{Input}{Input}\SetKwInOut{Output}{Output}
\Input{Matrix $\mathbf{Q}$, Vector $\mathbf{\Gamma}$ }
\Output{Vector q}
\BlankLine
A $\leftarrow$ zeros(4,4)\;
M $\leftarrow$ length($\mathbf{Q}$,1)\;
\BlankLine
\ForEach{i \text{in range(M)}}{
    $q$ = $\mathbf{Q}_i$\;
    $w_i$ = $\mathbf{\Gamma}_i$\;
    A = $w_i$ .* $(q * q')$ + A\; 
    alpha = alpha + $w_i$\;
}
A = (1.0 / alpha) * A\;
\BlankLine
\texttt{// Get the eigenvector corresponding to the largest eigenvalue}\;
q$\leftarrow$ eigs(A,1)\;
\end{algorithm}

\section{Spacecraft Pose Estimation Dataset (SPEED)}
Training a convolutional neural network usually requires extremely large labeled image datasets such as ImageNet\cite{Russakovsky2015} and Places\cite{Zhou2014}, which contain millions of images. Collecting and labeling such amount of actual space imagery is extremely difficult. Therefore, this work introduces SPEED, a dataset of images that enables not only training and validation of the SPN method but also benchmarking various state-of-the-art monocular vision-based pose estimation techniques. The SPEED images and the corresponding ground truth pose information are generated using two key complementary sources at the Space Rendezvous Laboratory (SLAB) of Stanford University\cite{slabURL}. The first source produced 15000 augmented reality images based on the Optical Stimulator \cite{Sharma2018a,Beierle2018} camera emulator software and actual images of the Earth captured by the Himawari-8 geostationary meteorological satellite \cite{Bessho2016}. The second source produced 300 actual camera images of a 1:1 mock-up of the Tango spacecraft using the Testbed for Rendezvous and Optical Navigation (TRON). Table~\ref{tab:camera} provides the camera model used for both these sources.
\begin{table*}
\renewcommand{\arraystretch}{1.3}
\caption{\bf Parameters of the camera used to capture the SPEED images.}
\label{tab:camera}
\centering
\begin{tabular}{|c|c|c|}
\hline
Parameter & Description & Value\\
\hline\hline
$N_u$ & Number of horizontal pixels & $1920$\\
$N_v$ & Number of vertical pixels & $1200$\\
$f_x$ & Horizontal focal length  & $0.0176$ m\\
$f_y$ & Vertical focal length  & $0.0176$ m\\
$du$ & Horizontal pixel length  & $5.86\cdot10^{-6}$ m\\
$dv$ & Vertical pixel length  & $5.86\cdot10^{-6}$ m\\
\hline
\end{tabular}
\end{table*}

\begin{figure}[htbp]
    \centering
    \begin{tabular}{ll}
        \includegraphics[trim={4cm 0cm 4.5cm 1cm}, clip, draft=false,width=0.35\linewidth]{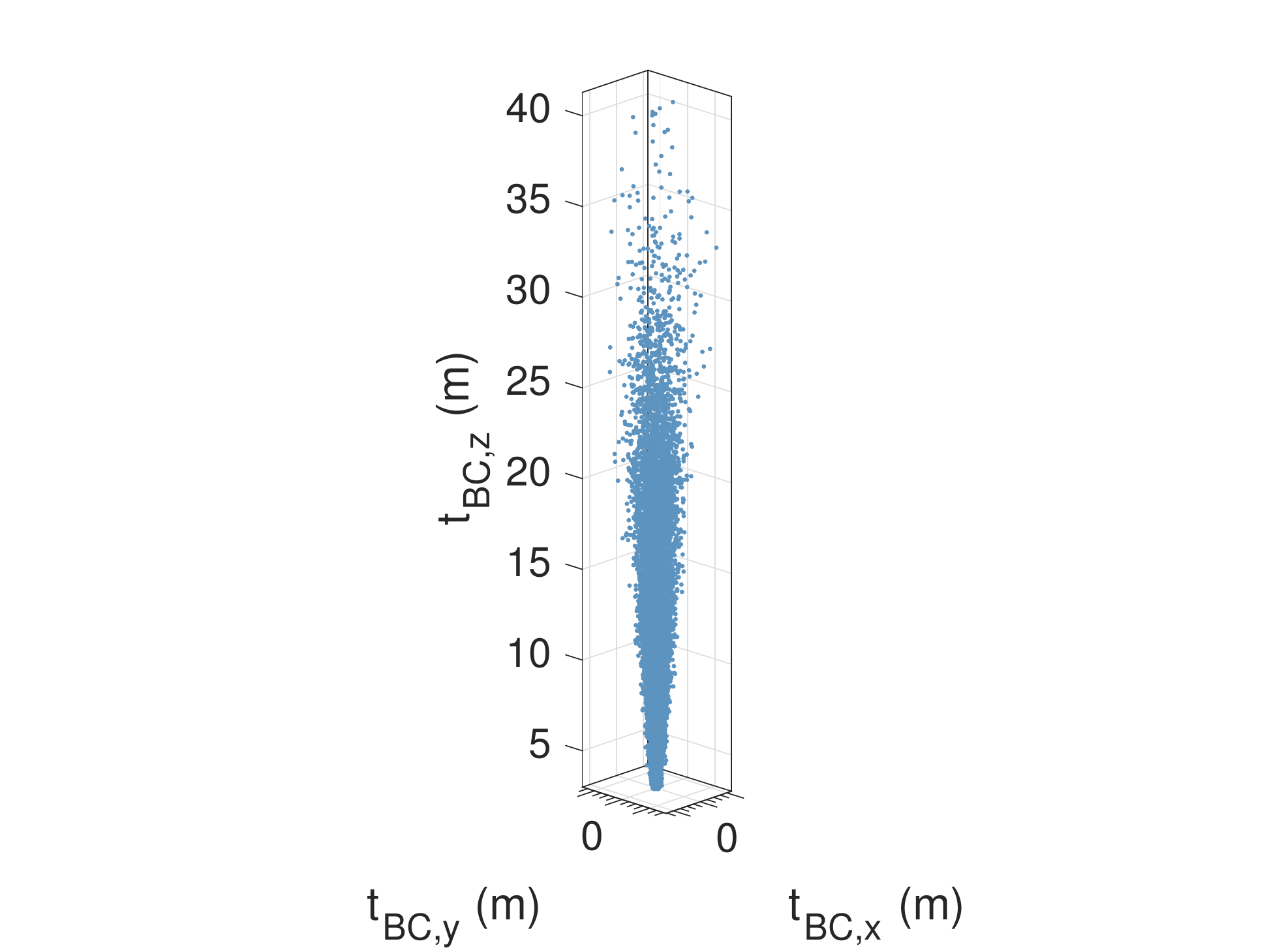} &
        \includegraphics[trim={1cm 0cm 3cm 1cm}, clip, draft=false,width=0.45\linewidth]{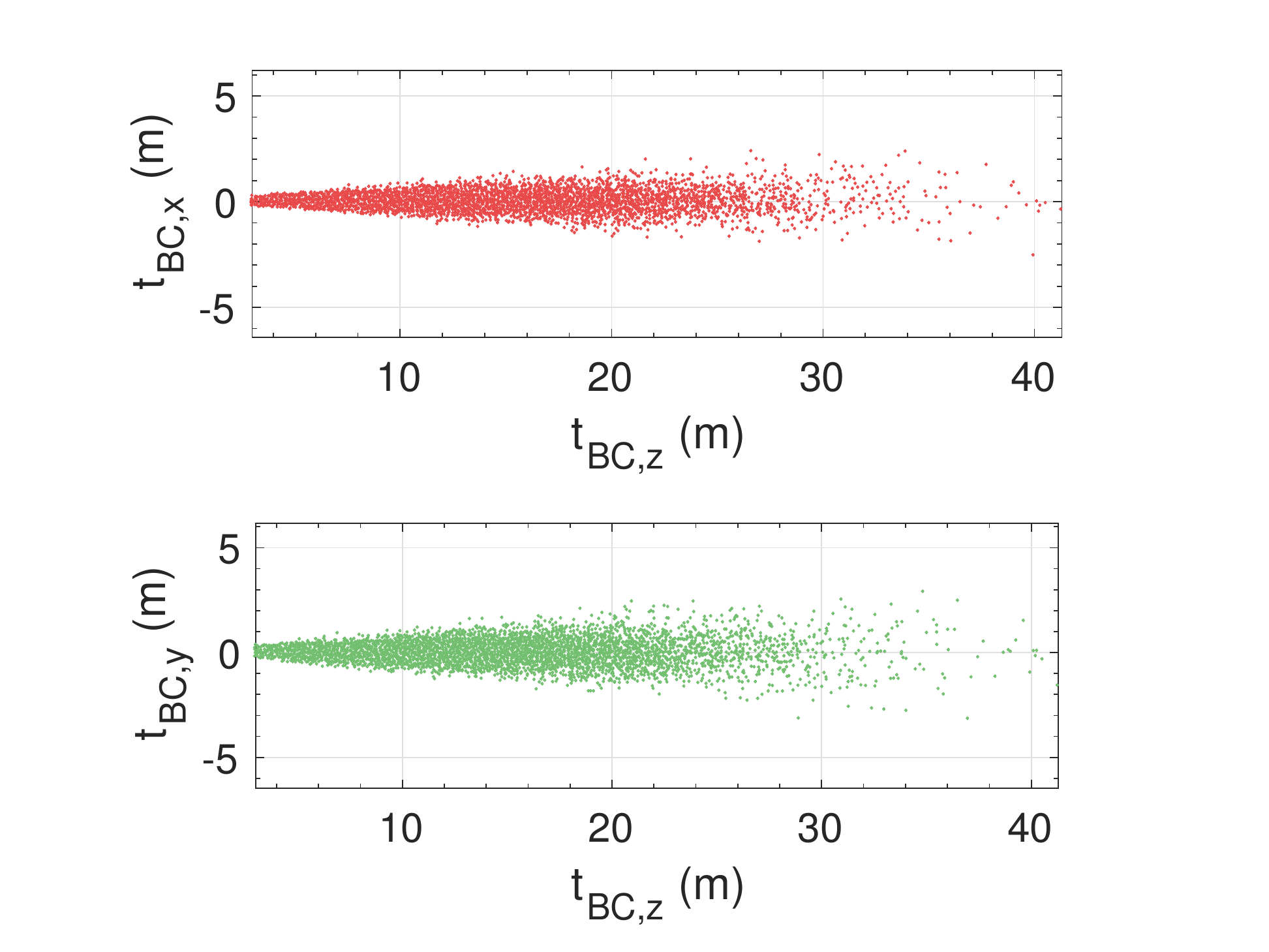} \\
    \end{tabular}
    \caption{The distribution of the relative position, $\mathbf{t_{BC}}$, in the SPEED images.}
    \label{fig:speedXyz}
\end{figure}

\begin{figure}[htbp]
    \centering
    \includegraphics[trim={3cm 0cm 3cm 0cm}, clip, width=0.75\textwidth]{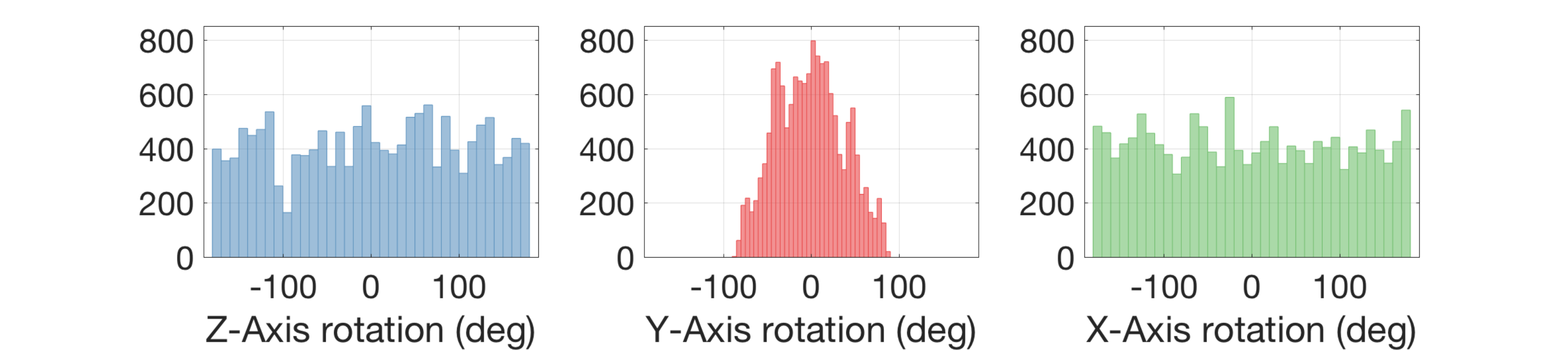}
    \caption{The distribution of the relative attitude in the SPEED images. For purposes of visualization, the relative attitude is parametrized as Euler angles.}
    \label{fig:speedAtt}
\end{figure}

The first source renders synthetic images of the Tango spacecraft using MATLAB and C++ language bindings of OpenGL. To create a diverse set of views of the target spacecraft, a set of relative attitudes and relative positions are selected. Unit quaternion parametrization of uniformly random rotations in the $SO(3)$ space are selected using Algorithm~\ref{alg:subgroup}. Figure~\ref{fig:speedAtt} shows the distribution of $\mathbf{q_{BC}}$ in the SPEED images, parametrized as Euler angles. The relative positions are obtained by separately selecting the relative distance and the bearing angles (defined in Equations~\ref{eq:bearingAngles1} and \ref{eq:bearingAngles2}). The bearing angles are uncorrelated random values selected from a multivariate normal distribution, $D_1 \sim \mathcal{N}(\bm{\mu} = [N_u/2,\; N_v/2]\text{px}, \;\bm{\sigma} = [5N_u/2,\; 5N_v/2]\text{px})$. The relative distance is randomly selected from a standard normal distribution, $D_2 \sim \mathcal{N}(\mu = 3\text{m}, \sigma = 10\text{m})$. Any relative distance values below 3 meters and above 50 meters are rejected. Figure~\ref{fig:speedXyz} shows the distribution of $\mathbf{t_{BC}}$ in the SPEED images.

\begin{figure}[htbp]
    \centering
    \includegraphics[width=0.95\textwidth]{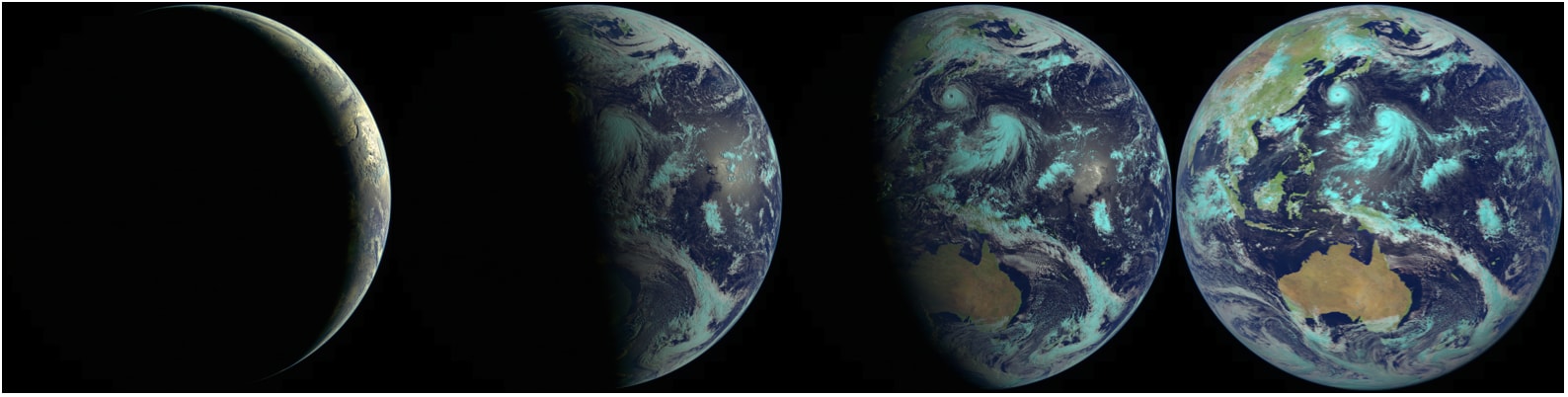}
    \caption{A montage of four of the 72 full-disk Earth images used to generate the background for half of the SPEED synthetic images.}
    \label{fig:earthImage}
\end{figure}

\begin{figure}[htbp]
    \centering
    \includegraphics[width=0.95\textwidth]{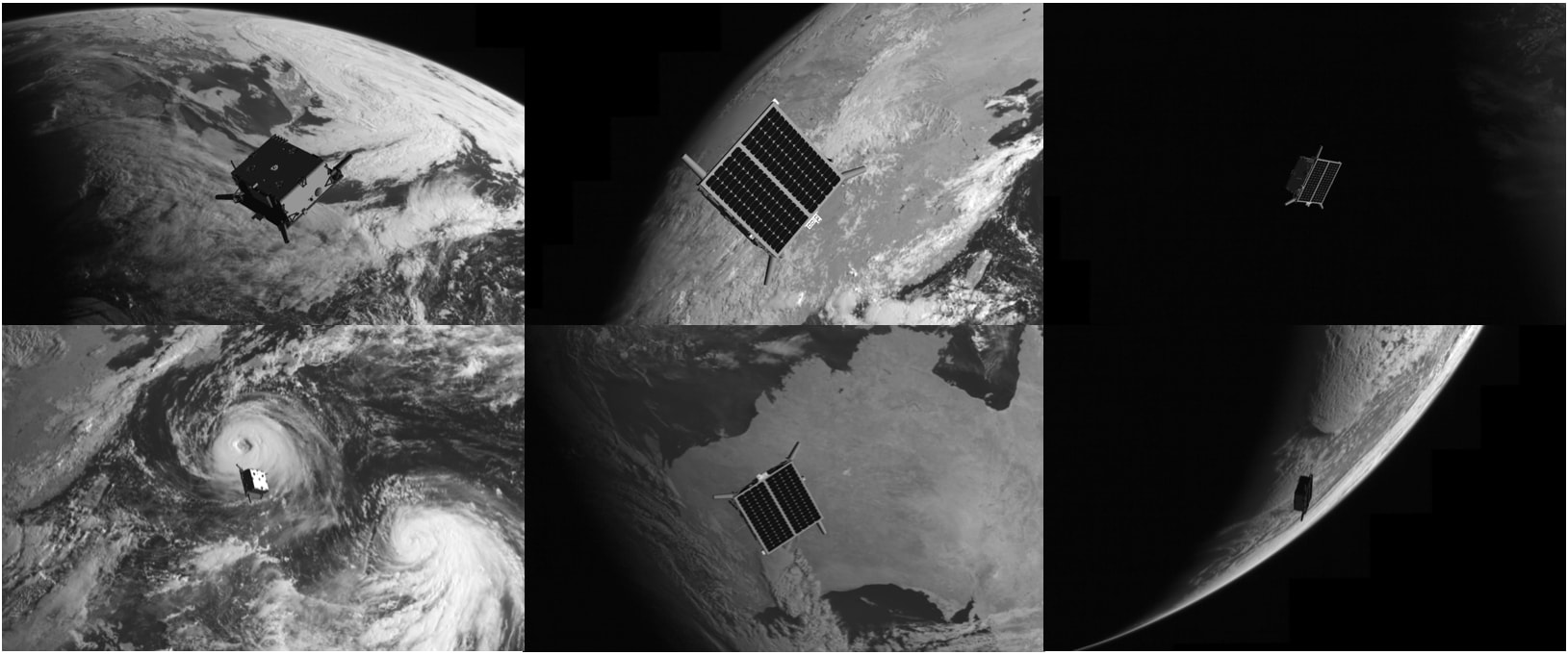}
    \caption{A montage of six synthetic images from the SPEED training-set.}
    \label{fig:speedSynthetic}
\end{figure}

The azimuth and elevation angles for the solar illumination are specifically chosen to match the solar illumination in the 72 actual images of the Earth captured by the Himawari-8 geostationary meteorological satellite. Figure~\ref{fig:earthImage} shows a montage of these images. The 72 images each provide a $100\cdot10^{6}$ pixels resolution disk-view of the Earth and were taken 10 minutes apart from each other over a period of 12 hours. Each of the images is converted to grayscale, cropped and inserted in as the background for half of the synthetic images of the Tango spacecraft. The location of the crop is selected at random from a uniform distribution spanning the Earth image. The size of the crop is selected to match the scale of the Earth when viewed through a camera (described in Table~\ref{tab:camera}) located at an altitude of 700 km and pointed in the nadir direction. Gaussian blurring and white noise are added to all images to emulate the depth of field and shot noise, respectively. Figure~\ref{fig:speedSynthetic} shows a montage of the resulting augmented reality images.

\begin{figure}[htbp]
    \centering
    \includegraphics[width=0.95\textwidth]{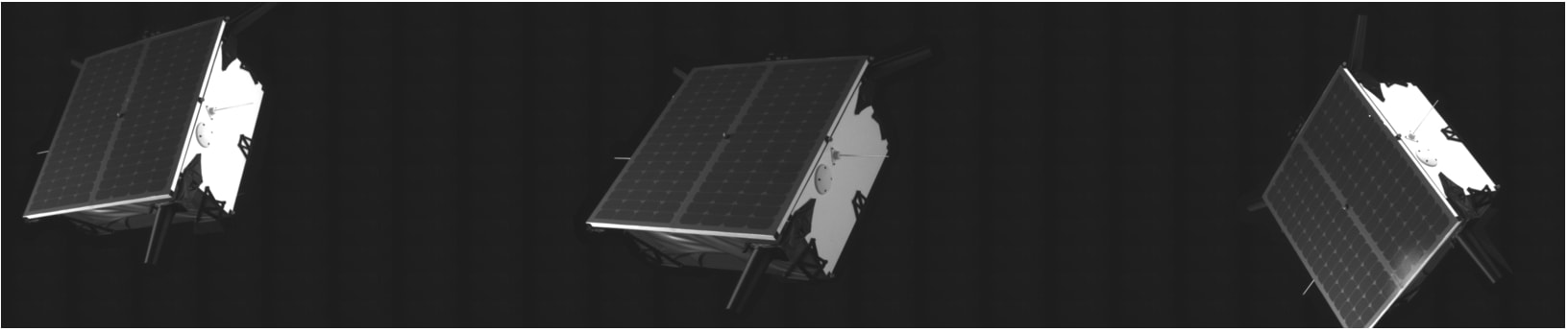}
    \caption{A montage of three actual camera images from the SPEED real test-set.}
    \label{fig:speedReal}
\end{figure}

The second source of the SPEED images is the TRON facility at SLAB. It consists of a 7 degrees-of-freedom robotic arm, which positions and orients a vision-based sensor with respect to a target object or scene. Custom illumination devices simulate Earth albedo and Sun light to high fidelity to emulate the illumination conditions present in space \cite{lightbox}. TRON provides images of a 1:1 mock-up model of the Tango spacecraft using a Point Grey Grasshopper 3 camera with a Xenoplan 1.9/17 mm lens. Note that this is the same camera as the one used in the first source. Calibrated motion capture cameras report the positions and attitudes of the camera and the Tango spacecraft, which are then used to calculate the “ground truth” pose of Tango with respect to the camera. Figure~\ref{fig:speedReal} shows a montage of the SPEED actual camera images. 

The SPEED training-set consists of 12000 synthetic images from the first source while the remainder 3000 synthetic images and the 300 actual camera images from the second source are available as two separate test-sets. The motivation for excluding the actual camera images from the training-set is to evaluate robustness and the domain adaptation capabilities of the pose estimation techniques.

\section{Experiments}
The proposed SPN method was trained and tested using the SPEED images. For all experiments, the region proposal network, the convolutional layers, and the fully connected layers were pre-trained on the relatively larger ImageNet dataset \cite{Russakovsky2015}. Following that, the region proposal network and the fully connected layers were trained using 80$\%$ of the SPEED training-set while the remainder 20$\%$ was used for validation. The two sets of fully connected layers were trained jointly while the region proposal network was trained separately. Both training routines were carried out using stochastic gradient in batches of 16 images. The initial learning-rate was set to 0.003 with an exponential decay of 0.95 every 1000 steps. For attitude determination, the hyper-parameters $m$ and $n$ were set to 1000 and 5, respectively. During training, each image was resized to 224 $\times$ 224 pixels to match the input size of the Zeiler and Fergus model architecture\cite{Zeiler2013}.

For quantitative analysis of the performance, three separate metrics are reported. To measure the accuracy of the 2D bounding box detection as compared with the ground truth 2D bounding box, the Intersection-Over-Union (IoU) metric is reported as
\begin{equation}
    \text{IoU} = \frac{\text{Area of Overlap}}{\text{Area of Union}}.
\end{equation}
To measure the accuracy of the estimated relative position, $\mathbf{\tilde{t}_{BC}}$ and the ground truth relative position, $\mathbf{t_{BC}}$, the translation error for each image can be calculated as
\begin{equation}
    E_T =  |\mathbf{t_{BC}} - \mathbf{\tilde{t}_{BC}}|
\end{equation}
To measure the accuracy of the estimated relative attitude, $\mathbf{\tilde{q}_{BC}}$ and the ground truth relative attitude, $\mathbf{q_{BC}}$, the attitude error for each image can be calculated as
\begin{align}
    E_R &=  2 \cos^{-1}(|z_s|),\text{ where} \\
    \mathbf{z} &= [z_s\;\mathbf{z_v}] = \mathbf{q_{BC}} * \text{conj}(\mathbf{\tilde{q}_{BC}}).
\end{align}

\begin{table*}[htbp]
\renewcommand{\arraystretch}{1.3}
\caption{\bf Performance of the SPN pose estimation method on the real and synthetic SPEED test-sets.}
\label{tab:accuracy}
\centering
\begin{tabular}{|c|c|c|}
\hline
Metric & SPEED synthetic test-set & SPEED real test-set\\
\hline\hline
Mean IoU (-) & 0.8582 & 0.8596\\
Median IoU (-) & 0.8908 & 0.8642\\
Mean $E_T$ (m) & [0.055 0.046 0.78]  & [0.036 0.015 0.189]\\
Median $E_T$ (m) & [0.024 0.021 0.496]  & [0.029 0.013 0.191]\\
Mean $E_R$ (deg) & 8.4254 & 18.188\\
Median $E_R$ (deg) & 7.0689 & 13.208\\
\hline
\end{tabular}
\end{table*}
Table~\ref{tab:accuracy} lists the mean and median values of the three performance metrics for the SPEED synthetic and real test-sets. The levels of the relative position and attitude errors improve upon those of other pose estimation methods applied on similar imagery in previous work\cite{Sharma2018,Sharma2018a}. Notably, the 2D bounding box detection is less affected by the gap between the synthetic and real test-sets as compared with the attitude estimation. The $E_T$ values for the actual camera images are lower than that of the synthetic images since the relative distances for the actual camera images is comparatively much lower than that of the synthetic images.

\begin{figure}[htbp]
    \centering    
    \includegraphics[width=0.95\textwidth]{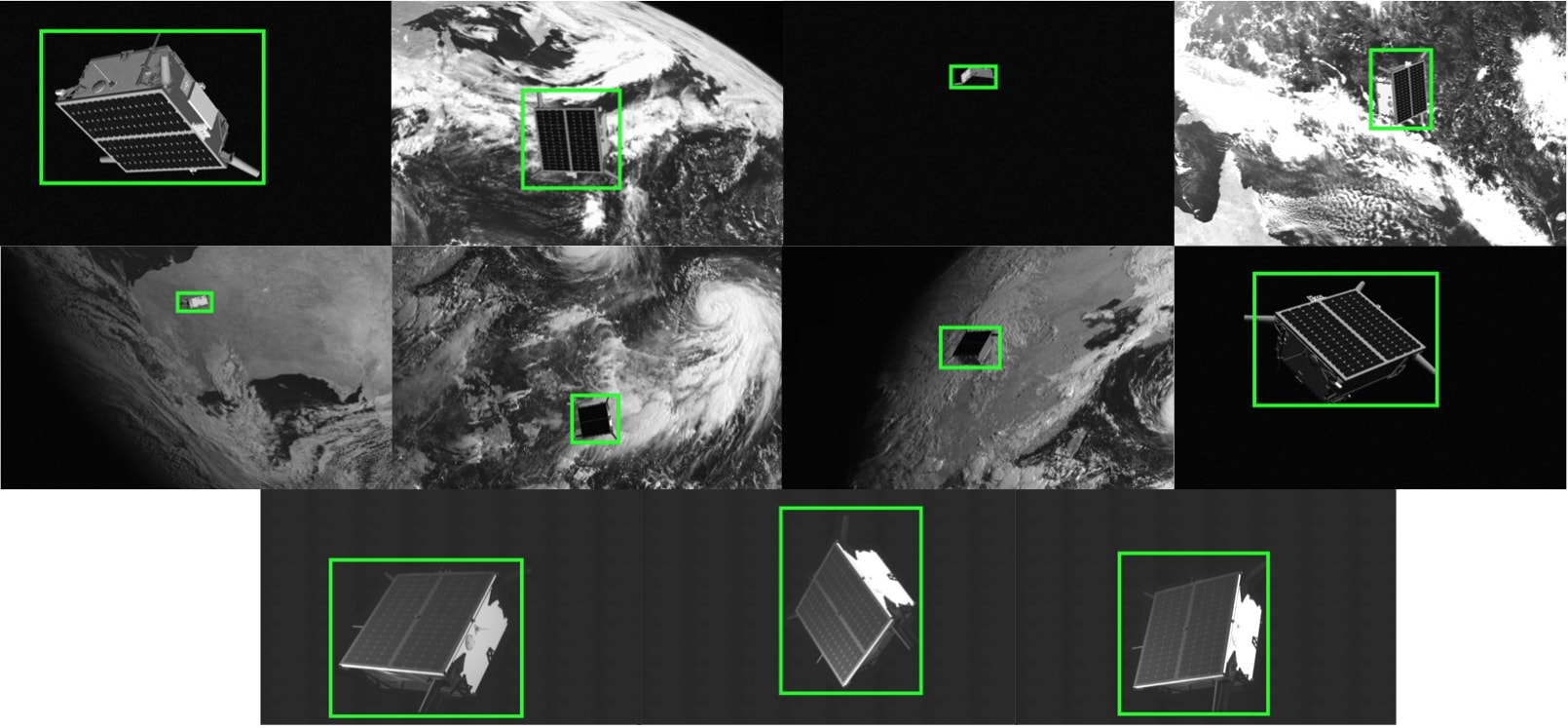}
    \caption{A montage of a few images with the 2D bounding box detections produced by the SPN method on the SPEED synthetic test-set.}
    \label{fig:bboxDetections}
\end{figure}
Figure~\ref{fig:bboxDetections} shows the qualitative results of the 2D bounding box detection on the SPEED synthetic and real test-sets. The region proposal network used in the SPN method was successful in detecting the 2D bounding box in images regardless of whether the Earth is visible in the background. In general, the bounding box detection worked marginally better on actual camera images as compared to synthetic images since all actual camera images had a black background. The bounding box detection was successful even in cases with strong shadows as long as one side of the spacecraft was illuminated.

\begin{figure}[htbp]
    \centering    
    \includegraphics[width=0.95\textwidth]{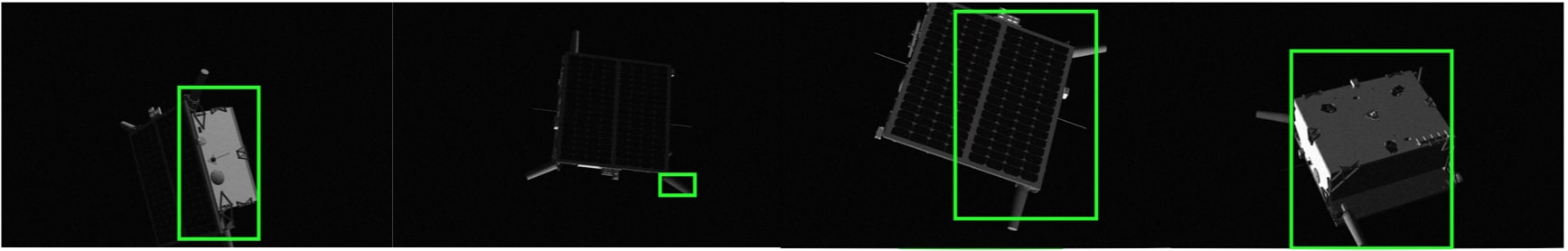}
    \caption{A montage of a few images from the SPEED synthetic test-set with inaccurate 2D bounding box detections.}
    \label{fig:bboxFails}
\end{figure}
However, the bounding box detection tends to fail in cases where the target spacecraft was in eclipse or when the target spacecraft was closer than $\sim$5 meters. Figure~\ref{fig:bboxFails} shows a few examples from both these failure modes. At relative distances less than $\sim$5 meters, the Tango spacecraft starts getting clipped by the image boundaries leading to unpredictable behavior of the detection. In a docking scenario, this could potentially be resolved by performing pose estimation relative to the docking mechanism or another smaller fixture, instead of the entire spacecraft. Note that the SPN method can be extended to detect multiple 2D bounding boxes corresponding to the fixtures of interest and performing attitude estimation and position determination for each.

To further examine such trends, the three performance metrics are plotted against the relative distance. In particular, the images from the SPEED synthetic test-set were grouped into batches of 100 each according to their ground truth relative distance, $||\mathbf{t_{BC}}||_2$. The mean value of the performance metric was then plotted against the mean ground truth relative distance for each batch.

\begin{figure}[htbp]
    \centering
    \includegraphics[width=0.7\textwidth]{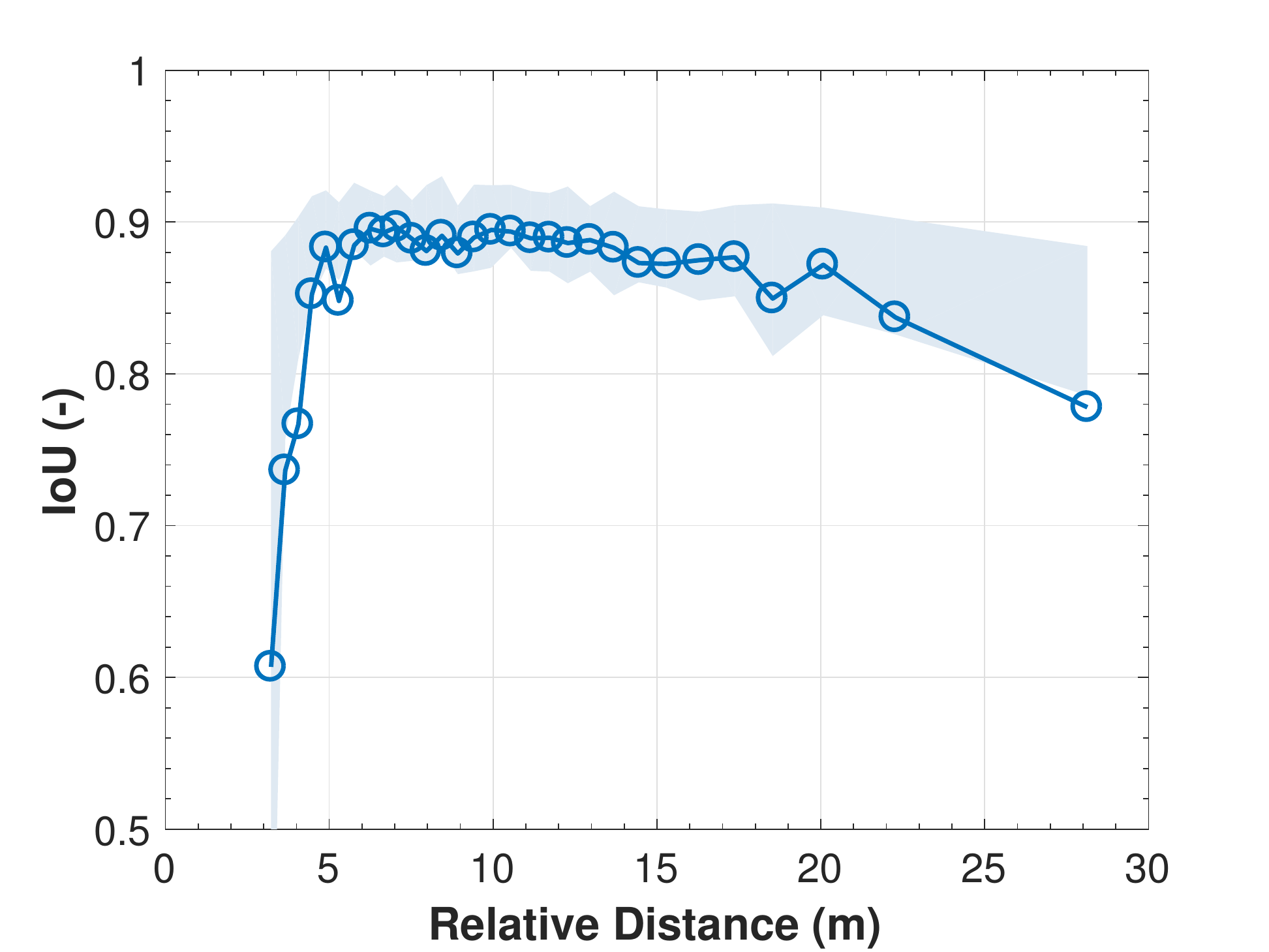}
    \caption{Mean IoU plotted against mean relative distance for the SPEED synthetic test-set. The shaded region shows the 25 and 75 percentile values.}
    \label{fig:iouVsRelDet}
\end{figure}

Figure~\ref{fig:iouVsRelDet} shows the mean IoU values of the SPN method for the SPEED synthetic test-set plotted against the mean relative distance. The bounding box detection has the highest accuracy at ranges of $\sim$7 meters to $\sim$20 meters while there is a sharp drop-off in bounding box detection at relatively closer distances of less than $\sim$5 meters. In contrast, there is a gradual drop-off in performance at relatively farther distances as the target spacecraft starts occupying too few pixels in the image plane to allow for an accurate 2D bounding box detection. Degradation in performance at larger separations is also contributed to by the fact that the SPEED training-set contains more images at lower relative distances.

\begin{figure}[htbp]
    \centering
    \includegraphics[width=0.7\textwidth]{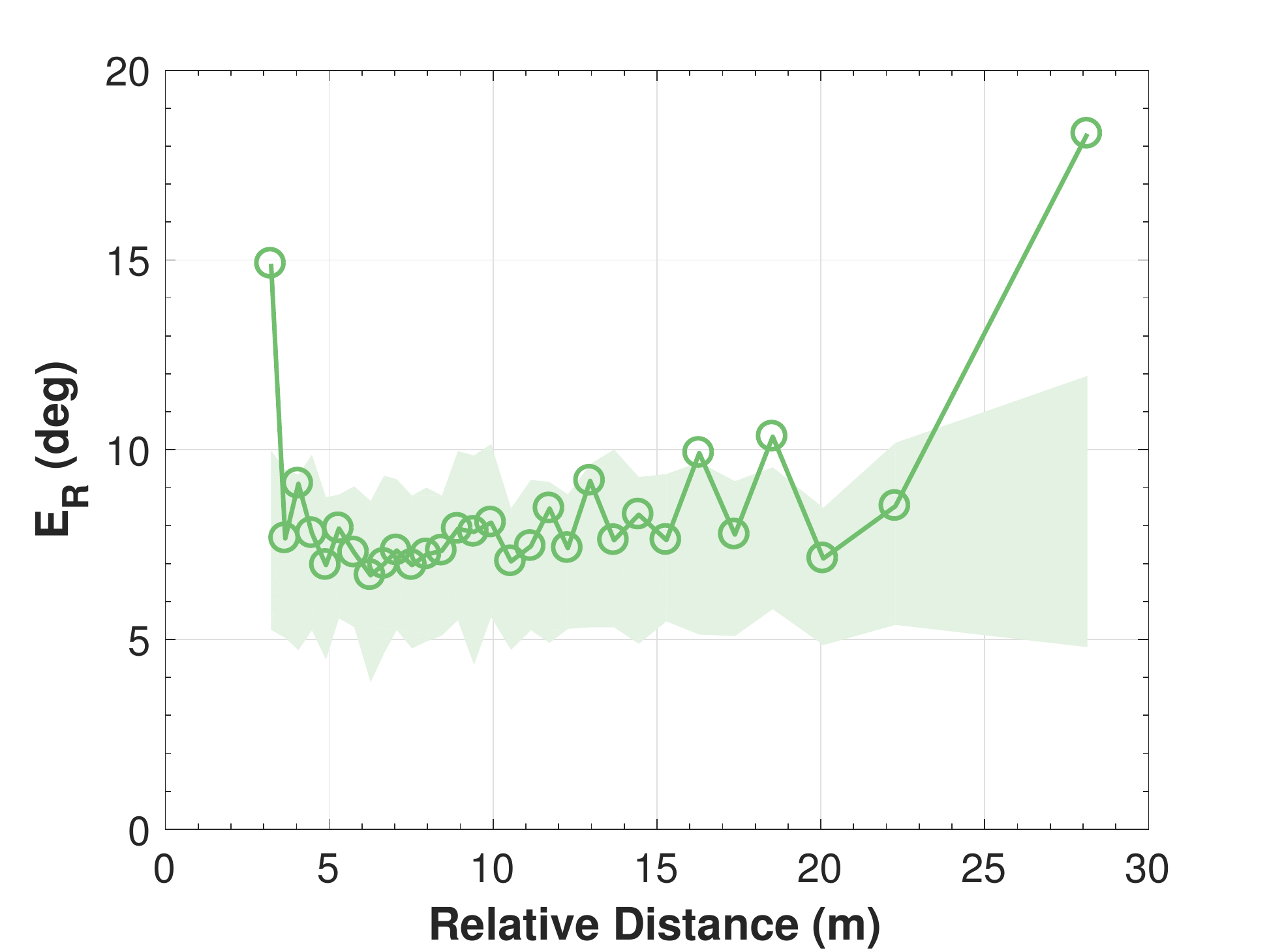}
    \caption{Mean $E_R$ plotted against mean relative distance for the SPEED synthetic test-set. The shaded region shows the 25 and 75 percentile values.}
    \label{fig:erVsRelDist}
\end{figure}
Figure ~\ref{fig:erVsRelDist} shows the mean $E_R$ values of the SPN method for the SPEED synthetic test-set plotted against the mean relative distance. The mean $E_R$ values are between 5 and 10 degrees for most of the relative distances. The range of the relative distances where attitude estimation has the highest accuracy is similar to the bounding box detection. This makes sense since both the region proposal network and the two sets of fully connected layers share the output of shared convolutional layers in the SPN method. Unlike the bounding box detection, the attitude estimation has a sharp drop-off in performance at both high and low relative distances. This is due to the presence of more outliers in the attitude estimation as compared with bounding box detection at high relative distances.

\begin{figure}[htbp]
    \centering
    \includegraphics[width=0.7\textwidth]{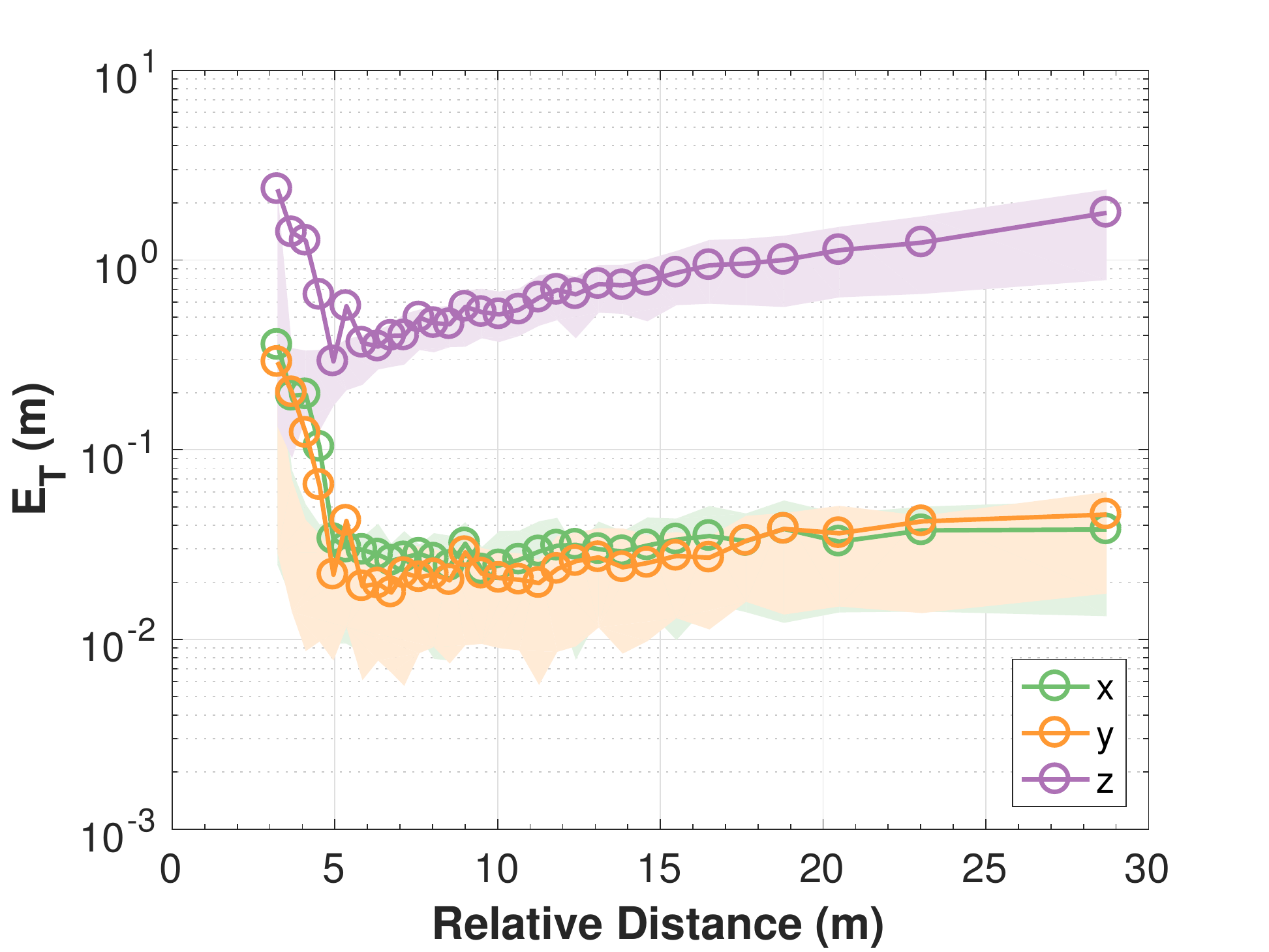} \\
    \caption{Mean $E_T$ plotted against mean relative distance for the SPEED synthetic test-set. The shaded region shows the 25 and 75 percentile values.}
    \label{fig:etVsRelDist}
\end{figure}
The estimated relative attitude value for each image in the SPEED synthetic test-set was then combined with the corresponding 2D bounding box detection to produce estimated values of the relative position. Figure~\ref{fig:etVsRelDist} shows the three components of mean $E_T$ values of the SPN method plotted against the mean relative distance. Note that the z-axis is aligned with the camera boresight direction while the x-axis and y-axis (lateral directions) are aligned with the image plane axes. The errors in the lateral directions was an order of magnitude lower than the camera boresight direction, which implies that the bounding box detection was fairly successful at estimating the center of the bounding box as compared with the size of the bounding box. Trends in performance degradation of $E_T$ mirror those observed for $E_R$ and IoU.

\begin{figure}[htbp]
    \centering
    \begin{tabular}{rl}
        \includegraphics[trim={4cm 0cm 4cm 1cm}, clip, draft=false,width=0.38\linewidth]{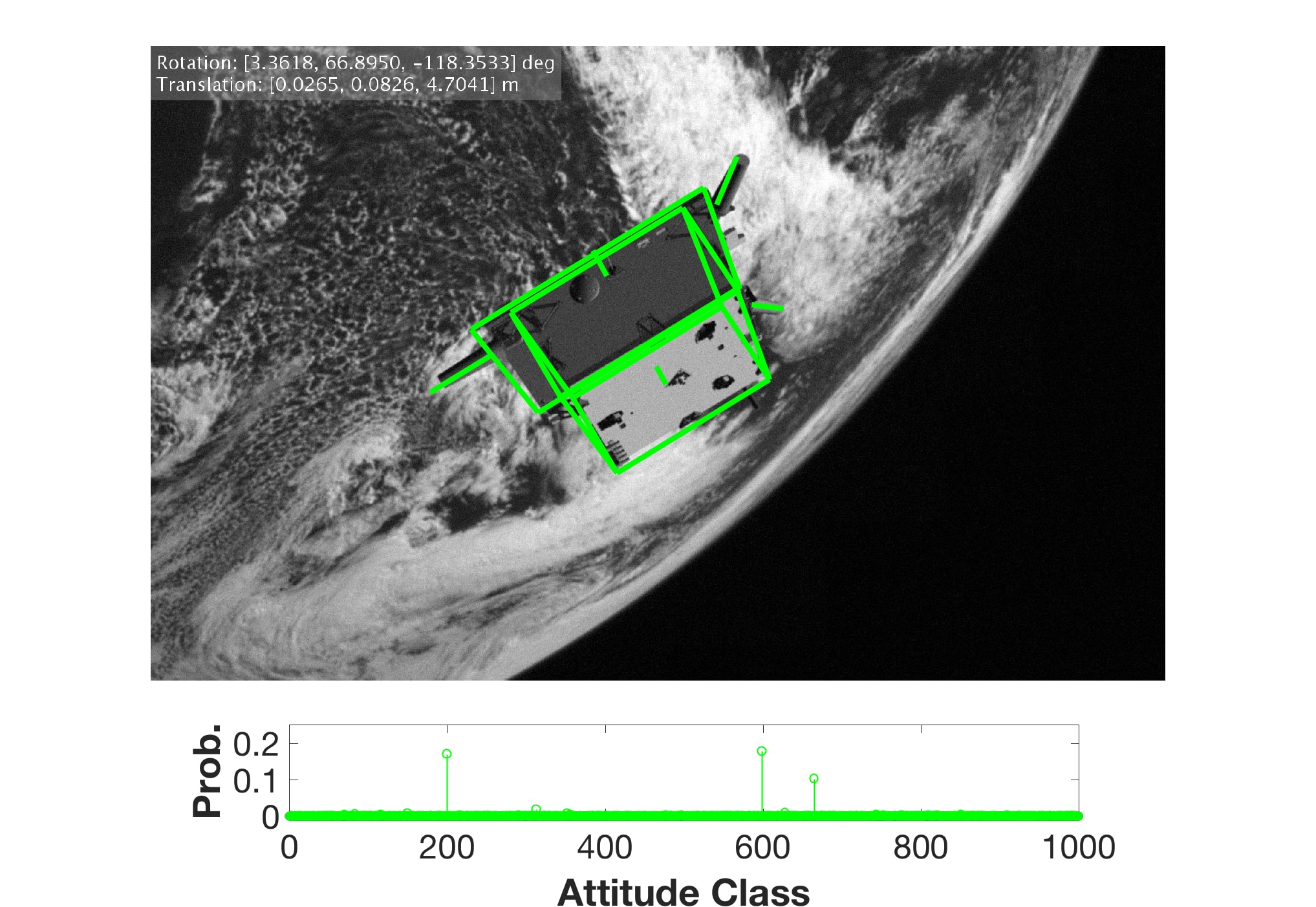} &    
        \includegraphics[trim={4cm 0cm 4cm 1cm}, clip, draft=false,width=0.38\linewidth]{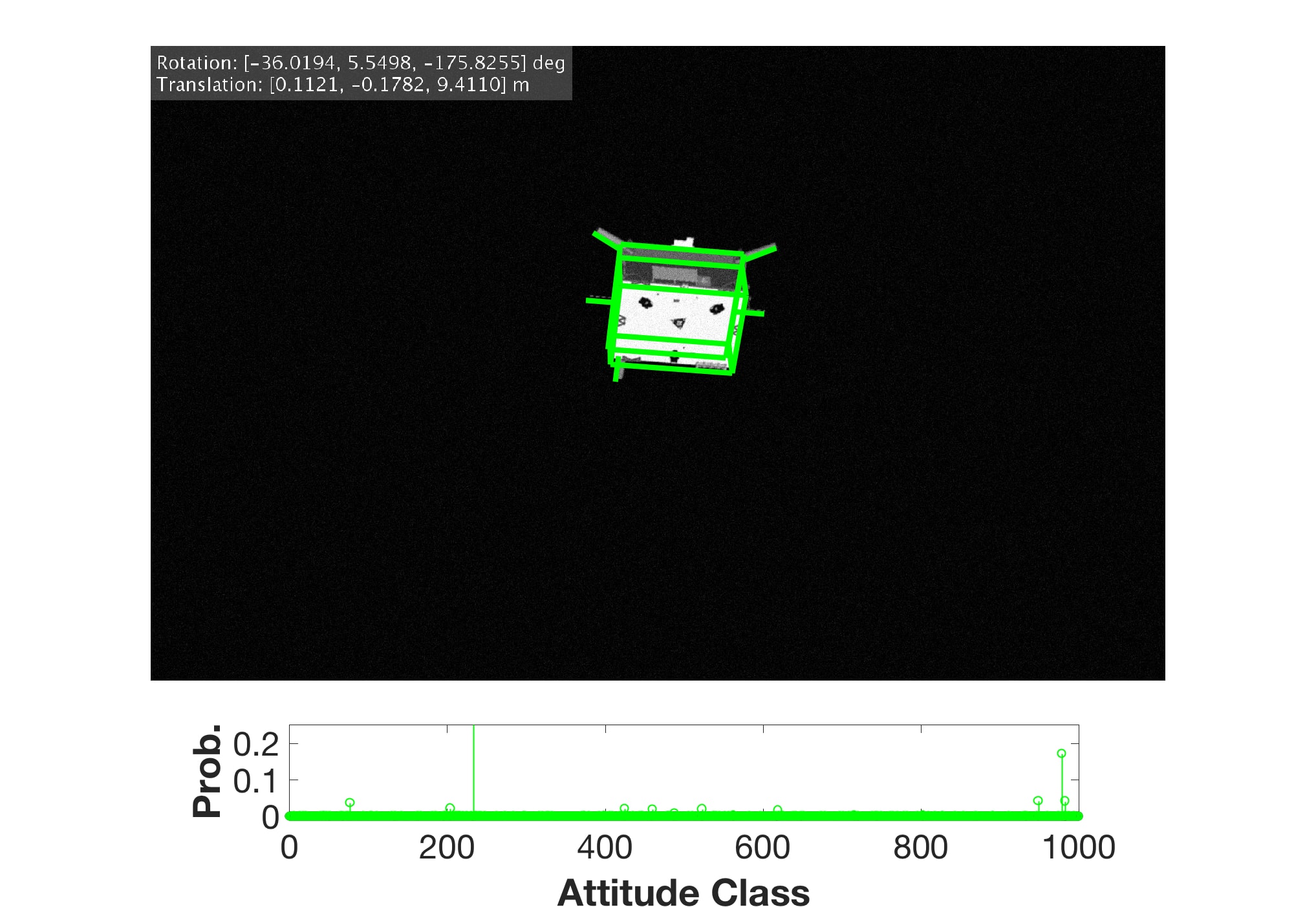} \\
        \includegraphics[trim={4cm 0cm 4cm 1cm}, clip, draft=false,width=0.38\linewidth]{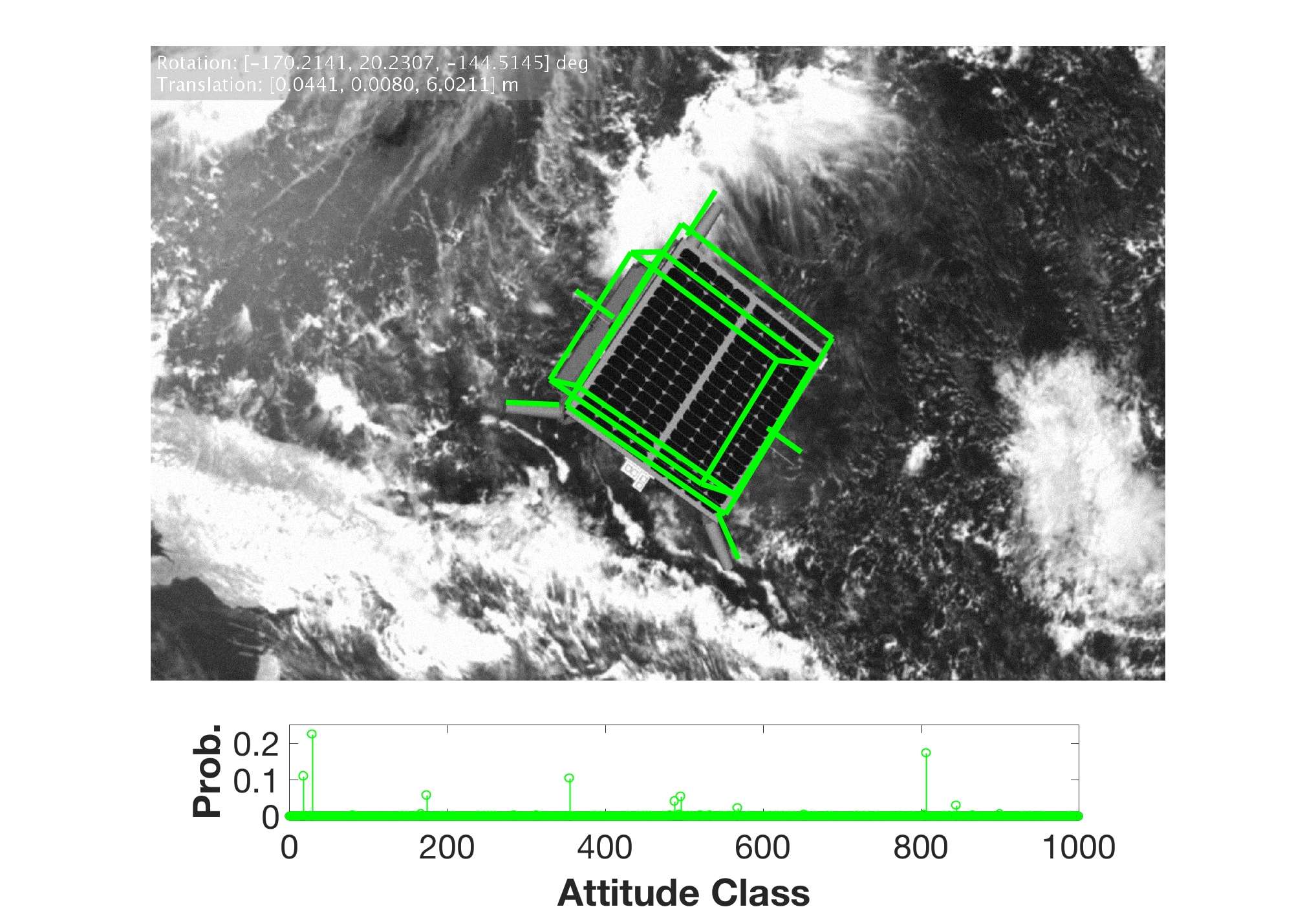} &    
        \includegraphics[trim={4cm 0cm 4cm 1cm}, clip, draft=false,width=0.38\linewidth]{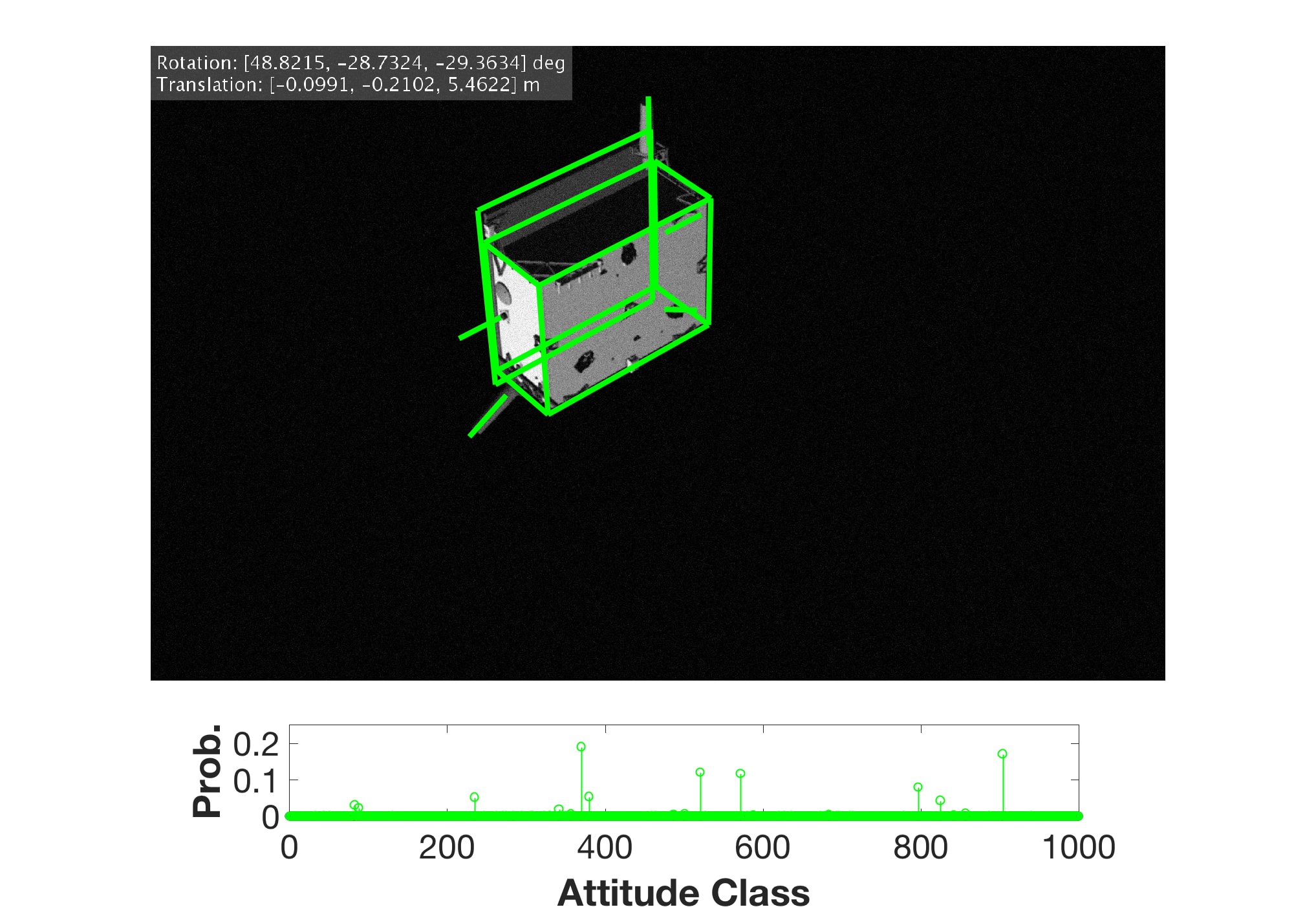} \\
    \end{tabular}
    \caption{A montage of a few images with the pose solutions produced by the SPN method on the on the SPEED synthetic test-set.} 
    \label{fig:poseSyn}
\end{figure}

\begin{figure}[htbp]
    \centering
    \begin{tabular}{rl}
        \includegraphics[trim={4cm 0cm 4cm 1cm}, clip, draft=false,width=0.38\linewidth]{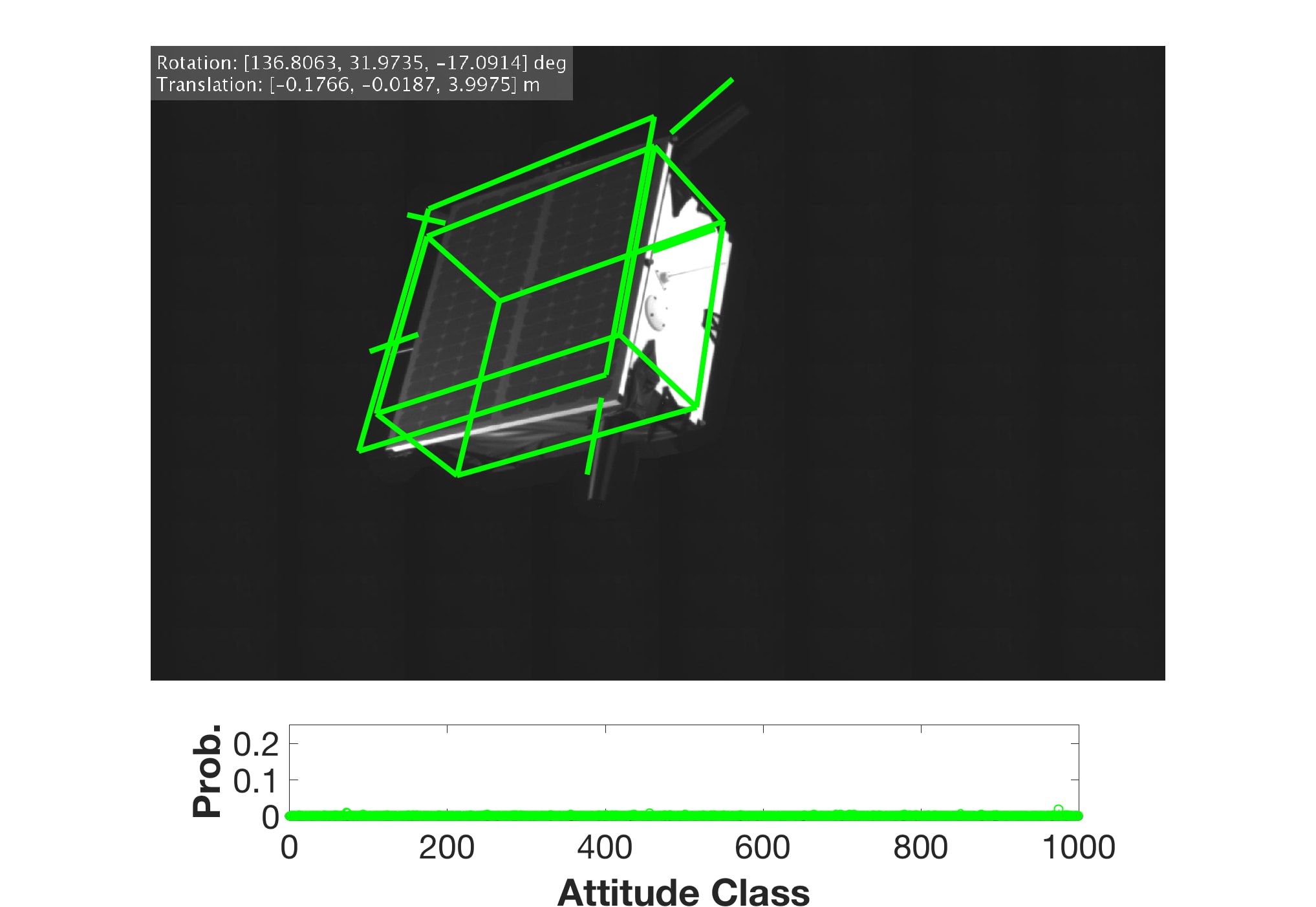} &    
        \includegraphics[trim={4cm 0cm 4cm 1cm}, clip, draft=false,width=0.38\linewidth]{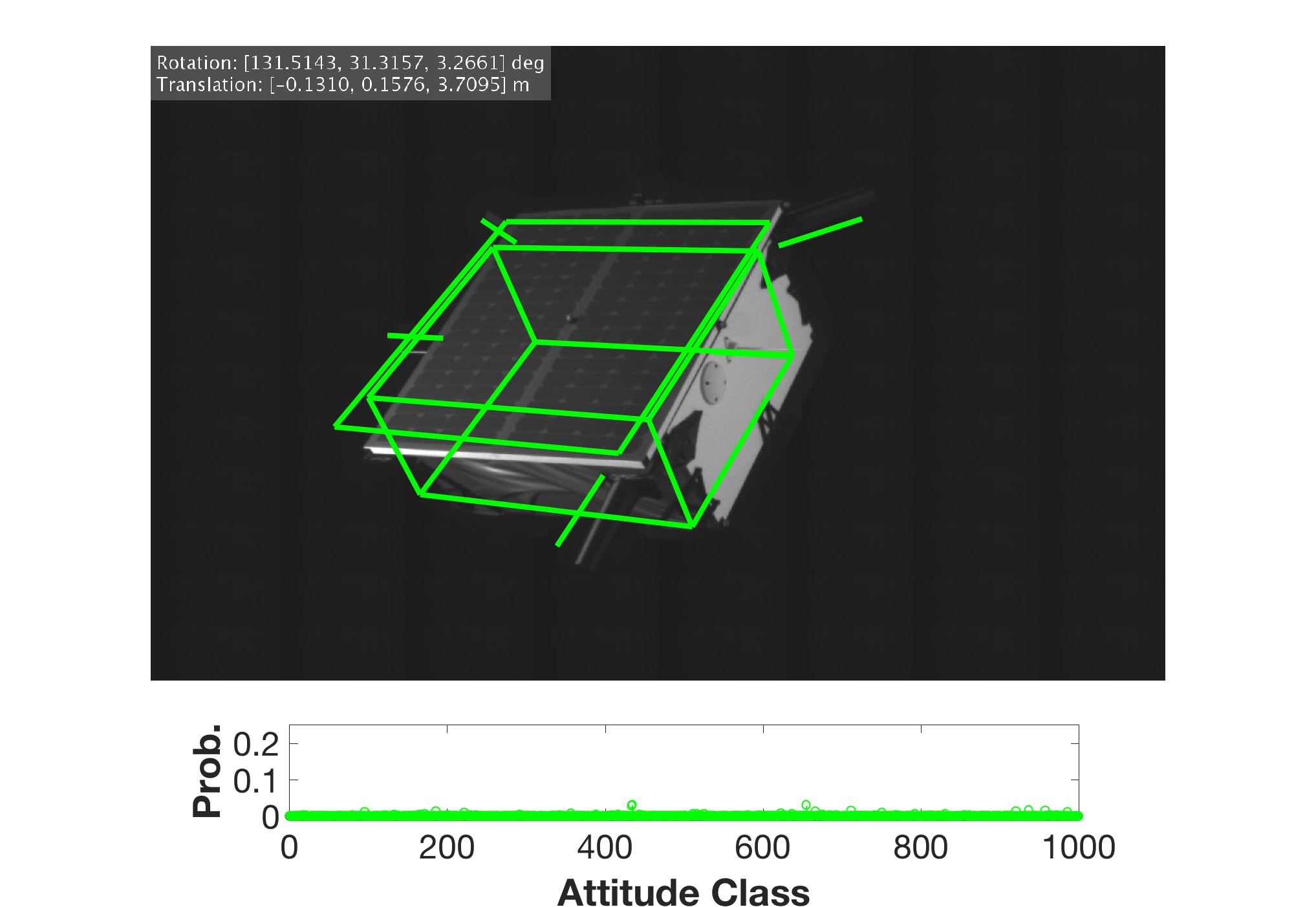} \\
    \end{tabular}
    \caption{A montage of a few images with the pose solutions produced by the SPN method on the on the SPEED real test-set.}
    \label{fig:poseReal}
\end{figure}

Figure~\ref{fig:poseSyn} and Figure~\ref{fig:poseReal} show some qualitative results of the pose estimated by the SPN method on the SPEED real and synthetic test-sets, respectively. The probability distribution output from the Branch 2 is also plotted for the respective images. In future, the probability distribution allows the setting up of a confidence metric to reject outliers in addition to being used by a navigation filter to accumulate information from sequential images to provide a more accurate estimate at a high rate. Predictably, the peaks in the probability distribution are lower for the SPEED real test-set as compared with the SPEED synthetic test-set since the convolutional neural network has overfit the synthetic images in the SPEED training-set to some extent. This could possibly be addressed by a stronger L2 regularization parameter and/or augmenting the training-set with images containing randomized textures of the target spacecraft similar to work in the area of domain adaptation\cite{Blitzer2007,Daume2009} and domain randomization\cite{OpenAI2018,Tobin2017}.


\section{Conclusions}
This work introduces the SPN method to estimate the relative pose of a target spacecraft using a single grayscale image without requiring a-priori pose information. The SPN method makes the novel use of a hybrid classification and regression technique to estimate the relative attitude. The SPN method leverages the geometric knowledge of the perspective transformation and advances in 2D bounding box detection to estimate the relative position using the Gauss-Newton algorithm. This work also introduces SPEED, a publicly available dataset to allow for the training and validation of monocular pose estimation techniques. The SPEED training-set consists of 12000 augmented reality images created by fusing synthetic images of a target spacecraft with actual camera images of the Earth. Apart from the 3000 images of the Tango spacecraft in the SPEED synthetic test-set, 300 actual camera images of the same spacecraft are provided in the SPEED real test-set. The subsequent application of the SPN method on the SPEED synthetic test-set produces degree-level mean relative attitude errors and centimeter-level mean relative position errors exceeding the performance of conventional feature-based methods used in previous work. The pose estimation performance carried over to the actual camera images as well albeit with slightly higher errors due to the gap between the synthetic images used during training and actual camera images used for testing. 

However, further work is required in a few directions. First, a complete assessment of how the SPN method stacks against conventional feature-based approaches as well as the more recent deep learning-based methods needs to be performed. The SPEED images and the associated performance metrics allow an excellent framework to carry out this assessment. Second, data augmentation techniques and/or stronger regularization during training is required to bridge the performance gap of the SPN method between the synthetic and real test-sets. Third, the performance drop-off at relative distances where the target spacecraft is only partially visible needs to be addressed to allow for pose estimation during all stages of close proximity operations. Lastly, the SPN method needs to be embedded in flight-grade hardware to profile its computational runtime and memory usage.

\section{Acknowledgments}
The authors would like to thank the King Abdulaziz City for Science and Technology (KACST) Center of Excellence for research in Aeronautics \& Astronautics (CEAA) at Stanford University for sponsoring this work. The authors would like to thank OHB Sweden, the German Aerospace Center (DLR), and the Technical University of Denmark (DTU) for the 3D model of the Tango spacecraft used to create the images used in this work. The authors would like to thank Nathan Stacey of the Space Rendezvous Laboratory at Stanford University for his technical contributions in generating ground truth pose information for the actual camera images of SPEED.

\bibliographystyle{AAS_publication}
\bibliography{library,urlbib}

\end{document}